\documentclass[journal=jacsat,manuscript=article]{achemso}

\usepackage{chemformula} 
\usepackage[T1]{fontenc} 
\usepackage{multirow}
\usepackage{booktabs}
\usepackage{amsmath, amssymb}

\author{Ziqi Zhang}
\affiliation[Jiangnan University]
{School of Artificial Intelligence and Computer Science, Jiangnan University}

\author{Shiheng Chen}
\affiliation[Jiangnan University]
{Key Laboratory of Industrial Biotechnology, Ministry of Education and School of Biotechnology and State Key Laboratory of Food Science and Resources, Jiangnan University}

\author{Runze Yang}
\affiliation[Macquarie University]
{School of Computing, Macquarie University}
\alsoaffiliation[Shanghai Jiao Tong University]
{Institute of Image Processing and Pattern Recognition, Shanghai Jiao Tong University, and Key Laboratory of System Control and Information Processing, Ministry of Education of China}

\author{Zhisheng Wei}
\affiliation[Jiangnan University]
{Key Laboratory of Industrial Biotechnology, Ministry of Education and School of Biotechnology and State Key Laboratory of Food Science and Resources, Jiangnan University}

\author{Wei Zhang}
\affiliation[Nantong University]
{School of Artificial Intelligence and Computer Science, Nantong University}

\author{Lei Wang}
\affiliation[Jiangnan University]
{Key Laboratory of Industrial Biotechnology, Ministry of Education and School of Biotechnology and State Key Laboratory of Food Science and Resources, Jiangnan University}

\author{Zhanzhi Liu}
\affiliation[Jiangnan University]
{Key Laboratory of Industrial Biotechnology, Ministry of Education and School of Biotechnology and State Key Laboratory of Food Science and Resources, Jiangnan University}
\alsoaffiliation[]{Shandong Huatai Paper Co. Ltd and Shandong Yellow Triangle Biotechnology Industry Research Institute Co. Ltd}

\author{Fengshan Zhang}
\affiliation[]{Shandong Huatai Paper Co. Ltd and Shandong Yellow Triangle Biotechnology Industry Research Institute Co. Ltd}

\author{Jing Wu}
\affiliation[Jiangnan University]
{Key Laboratory of Industrial Biotechnology, Ministry of Education and School of Biotechnology and State Key Laboratory of Food Science and Resources, Jiangnan University}
\alsoaffiliation[]{Shandong Huatai Paper Co. Ltd and Shandong Yellow Triangle Biotechnology Industry Research Institute Co. Ltd}

\author{Xiaoyong Pan}
\affiliation[Shanghai Jiao Tong University]
{Institute of Image Processing and Pattern Recognition, Shanghai Jiao Tong University, and Key Laboratory of System Control and Information Processing, Ministry of Education of China}

\author{Hongbin Shen}
\affiliation[Shanghai Jiao Tong University]
{Institute of Image Processing and Pattern Recognition, Shanghai Jiao Tong University, and Key Laboratory of System Control and Information Processing, Ministry of Education of China}

\author{Longbing Cao}
\affiliation[Macquarie University]
{School of Computing, Macquarie University}

\author{Zhaohong Deng}
\affiliation[Jiangnan University]
{School of Artificial Intelligence and Computer Science, Jiangnan University}
\email{dengzhaohong@jiangnan.edu.cn}

\title[An \textsf{achemso} demo]
  {Modeling enzyme temperature stability from sequence segment perspective}


\begin{document}

\begin{tocentry}
    \centering
    \includegraphics[width=0.75\linewidth]{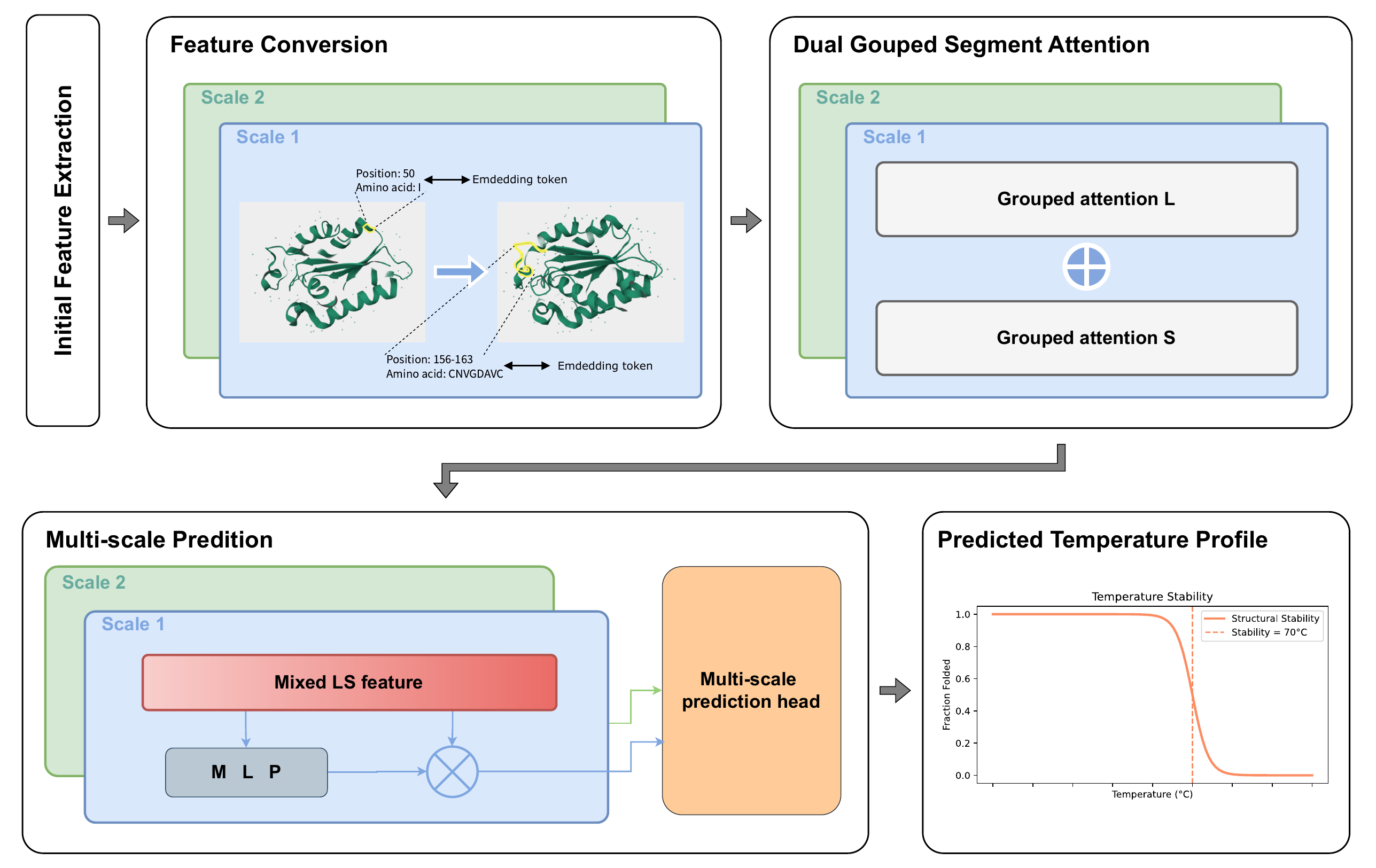}
\end{tocentry}


\begin{abstract}
  Developing enzymes with desired thermal properties is crucial for a wide range of industrial and research applications, and determining temperature stability is an essential step in this process. Experimental determination of thermal parameters is labor-intensive, time-consuming, and costly. Moreover, existing computational approaches are often hindered by limited data availability and imbalanced distributions. To address these challenges, we introduce a curated temperature stability dataset designed for model development and benchmarking in enzyme thermal modeling. Leveraging this dataset, we present the \textit{Segment Transformer}, a novel deep learning framework that enables efficient and accurate prediction of enzyme temperature stability. The model achieves state-of-the-art performance with an RMSE of 24.03, MAE of 18.09, and Pearson and Spearman correlations of 0.33, respectively. These results highlight the effectiveness of incorporating segment-level representations, grounded in the biological observation that different regions of a protein sequence contribute unequally to thermal behavior. As a proof of concept, we applied the Segment Transformer to guide the engineering of a cutinase enzyme. Experimental validation demonstrated a 1.64-fold improvement in relative activity following heat treatment, achieved through only 17 mutations and without compromising catalytic function.
\end{abstract}

\section{Introduction}
Temperature stability is among the most critical factors determining an enzyme's suitability for industrial and research applications \cite{daniel2001temperature, daniel2008effect, daniel2010new, lee2013determination}. It reflects an enzyme's ability to maintain structural integrity and catalytic function under thermal stress. To mine or design enzymes that function reliably under specific temperature conditions, three principal methodologies are commonly employed: directed evolution, rational design, and semi-rational design. All these approaches require accurate assessment of temperature stability \cite{xu2020recent}. However, experimental determination is labor-intensive, time-consuming, and costly.

Although machine learning (ML) has been widely adopted in protein engineering, recent progress in modeling enzyme temperature stability remains limited. A key challenge is the scarcity and imbalance of annotated data. Despite the existence of over 230 million enzyme sequences in the UniProt database, fewer than 30,000 experimentally determined temperature stability records are available in the BRENDA database \cite{10.1093/nar/gkae1010, 10.1093/nar/gkaa1025}. This disparity presents a substantial barrier to data-driven model development. Additionally, data imbalance—especially the underrepresentation of thermostable enzymes—limits current models’ ability to generalize across the full temperature spectrum.

To address these challenges, we curated a comprehensive enzyme temperature stability dataset from BRENDA and performed a detailed analysis of its distribution. To enhance generalization and robustness, we implemented a data partitioning strategy that minimizes sequence similarity between the training and validation sets. This curated dataset provides a reliable benchmark for modeling enzyme thermal properties. To mitigate the effects of data imbalance, we employed a weighted RMSE loss function that emphasizes accurate predictions for underrepresented enzymes in extreme temperature ranges.

Existing computational methods often suffer from narrow applicability. Several models are tailored to specific enzyme classes \cite{yan2012prediction, yan2019predictors, zhang2012prediction, chu2016predicting, foroozandeh2021generalized}, limiting their generalizability. Others require organism-dependent features such as the optimal growth temperature (OGT) of the source organism \cite{li2019machine, gado2020improving, wang2024artificial}, which restricts their use when such metadata are unavailable. While recent OGT-independent models have improved accessibility \cite{li2022learning, zhang2022novel, qiu2024seq2topt}, their performance remains suboptimal for practical enzyme engineering tasks.

A fundamental limitation of existing approaches is the neglect of heterogeneous contributions across different regions of enzyme sequences to thermal behavior. Segment Transformer addresses this gap by introducing \emph{segment-level features}, in contrast to conventional residue-level embeddings where each amino acid corresponds to a single token \cite{9477085, lin2022language}. Segment-level representations aggregate information from short, contiguous sequence segments, aligning with biological insights that different regions of a protein contribute unequally to thermostability \cite{chronopoulou2023key, lee2014protein, kannan2000aromatic}.

We thus propose \textit{Segment Transformer}, a user-friendly, high-accuracy computational model for predicting enzyme temperature stability. The model outputs not only point estimates but also its fluctuation ranges, providing interpretable and actionable predictions. It achieves an RMSE of 24.03, MAE of 18.09, and Pearson and Spearman correlations of 0.33, outperforming existing tools. More importantly, Segment Transformer offers a new optimization perspective distinct from traditional $\Delta\Delta G$-based or structure-based methods \cite{musil2017fireprot, liu2006rosettadesign, goldenzweig2016automated}.

In an application study, Segment Transformer successfully guided the thermostability enhancement of a cutinase from \textit{Humicola insolens} (HiC). The engineered variant demonstrated a 1.64-fold increase in relative activity and a 3.9-fold increase in half-life after heat treatment, achieved with only 17 mutations and no reduction in activity. These results highlight the potential of Segment Transformer as a practical and effective tool in computational enzyme engineering.

\section{Methodology}

\subsection{Overview of Workflow}

\begin{figure}
    \centering
    \includegraphics[width=1\linewidth]{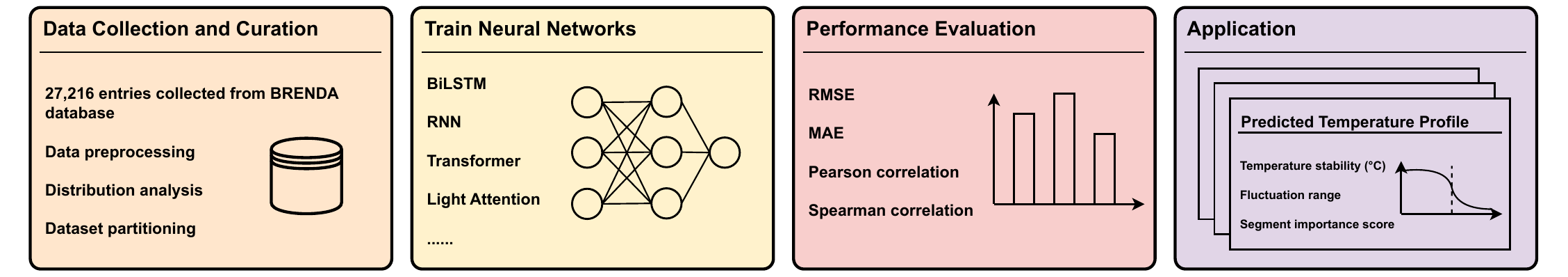}
    \caption{Overall research workflow for developing and validating the Segment Transformer.}
    \label{fig:workflow}
\end{figure}

In this study, we present the \textit{Segment Transformer}, a novel deep learning framework that leverages segment-level representations to model the relationship between enzyme sequences and their temperature stability. As illustrated in Fig.~\ref{fig:workflow}, the development process follows a structured and multi-stage workflow.

The first stage involves data collection and curation. One of the primary bottlenecks in applying large-scale machine learning to enzyme temperature modeling is the lack of sufficiently large and diverse training datasets. To address this, we collected temperature stability records from the BRENDA enzyme functional database as of December 2024, comprising 27,216 curated entries. We then performed preprocessing, distributional analysis, and designed a partitioning strategy that minimizes sequence similarity between subsets to ensure reliable model evaluation. Further details are provided in the section \textit{Data Collection and Curation}.

In the second stage, we benchmarked several neural network architectures to identify effective modeling strategies. Through extensive experiments, we observed that incorporating segment-level features—where each embedding token represents a short sequence fragment—markedly improves predictive performance. This insight led to the design and implementation of the Segment Transformer architecture.

The third stage involves rigorous performance evaluation of the Segment Transformer against existing methods. Detailed comparisons, including multiple regression metrics and correlation analyses, are provided in the section \textit{Performance Evaluation}.

While computational simulation is essential, it does not fully capture the practical utility of a predictive model. Therefore, in the fourth stage, we applied the Segment Transformer to guide the thermostability engineering of a cutinase enzyme. Experimental results validating this application are described in the section \textit{Application Study of Segment Transformer}.

\subsection{Data Collection and Curation}\label{sec_data}

\begin{figure}
    \centering
    \includegraphics[width=1\linewidth]{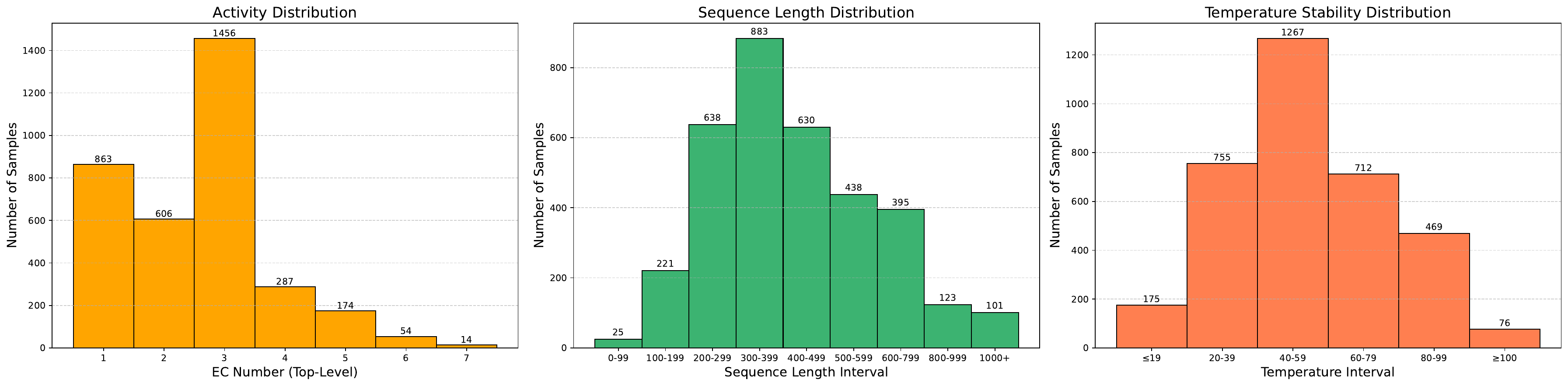}
    \caption{Distribution analysis of the curated dataset, including EC classification, sequence length, and temperature stability.}
    \label{fig:distribution}
\end{figure}

Temperature stability data were retrieved from the BRENDA enzyme functional database in December 2024, resulting in an initial collection of 27,216 entries. The following preprocessing steps were applied:

\begin{enumerate}
    \item Samples without UniProt accession numbers or missing numerical thermal stability annotations were removed.
    \item Samples containing multiple accession numbers were split such that each entry contained a single accession number paired with one thermal parameter.
    \item Duplicate entries based on accession numbers were discarded.
\end{enumerate}

After preprocessing, the final dataset consisted of 3,454 unique entries. We performed a distributional analysis of the data from three perspectives: activity class (EC number), sequence length, and temperature stability (Fig.~\ref{fig:distribution}). EC class 3 (hydrolases) contained the largest number of samples (1,456), followed by EC class 1 (oxidoreductases). In terms of sequence length, 86.39\% of the samples fell within the 200–799 residue range. For temperature stability, the distribution was highly imbalanced: 36.68\% of samples (1,267 entries) were concentrated between 40°C and 59°C, while extreme temperatures were underrepresented—only 175 samples at $\leq$19°C and 76 at $\geq$100°C. This skewed distribution presents a challenge for training generalizable models.

\begin{table}
    \centering
    \begin{tabular}{ccc|ccc}
    \toprule
         \multicolumn{3}{c|}{\textbf{Grouping}} & \multicolumn{3}{c}{\textbf{Partition}}\\
    \midrule
         \textbf{Temperature Range} & \textbf{Sequences} & \textbf{Clusters} & \textbf{Train} & \textbf{Validation} & \textbf{Test}\\
         $t < 45^\circ$C  & 1,237  & 678 & \multirow{4}{*}{2,798} & \multirow{4}{*}{311} & \multirow{4}{*}{345}\\
         $45 \leq t < 70^\circ$C & 1,136 & 606 &  &  & \\
         $70 \leq t < 100^\circ$C & 665 & 422 &  &  & \\
         $t \geq 100^\circ$C & 71 & 66 &  &  & \\
    \bottomrule
    \end{tabular}
    \caption{Dataset statistics by temperature range and final partitioning.}
    \label{tab:data}
\end{table}

Given the high cost of wet-lab experiments, data scarcity remains a major obstacle in enzyme thermal modeling. To make efficient use of the available data, we adapted and extended the partitioning strategy proposed in literature\cite{gado2025machine}. Specifically, 10\% of the full dataset was randomly assigned as a held-out test set. Partitioning strategies were applied only to the remaining data (train + validation), with the test set kept fixed across all experiments.

For the training and validation sets, sequences were first grouped by temperature intervals. Within each group, similar sequences were clustered using MMseqs2 \cite{steinegger2017mmseqs2}. Then, 90\% of clusters were randomly assigned to the training set and 10\% to the validation set. This strategy minimizes sequence identity overlap between training and validation data, encouraging models to generalize to structurally dissimilar inputs. The final distribution of sequences and clusters across partitions is summarized in Table~\ref{tab:data}.

\subsection{Framework Overview}\label{sec_segment_transformer}

\begin{figure}[htbp]
\centering
\includegraphics[width=\linewidth]{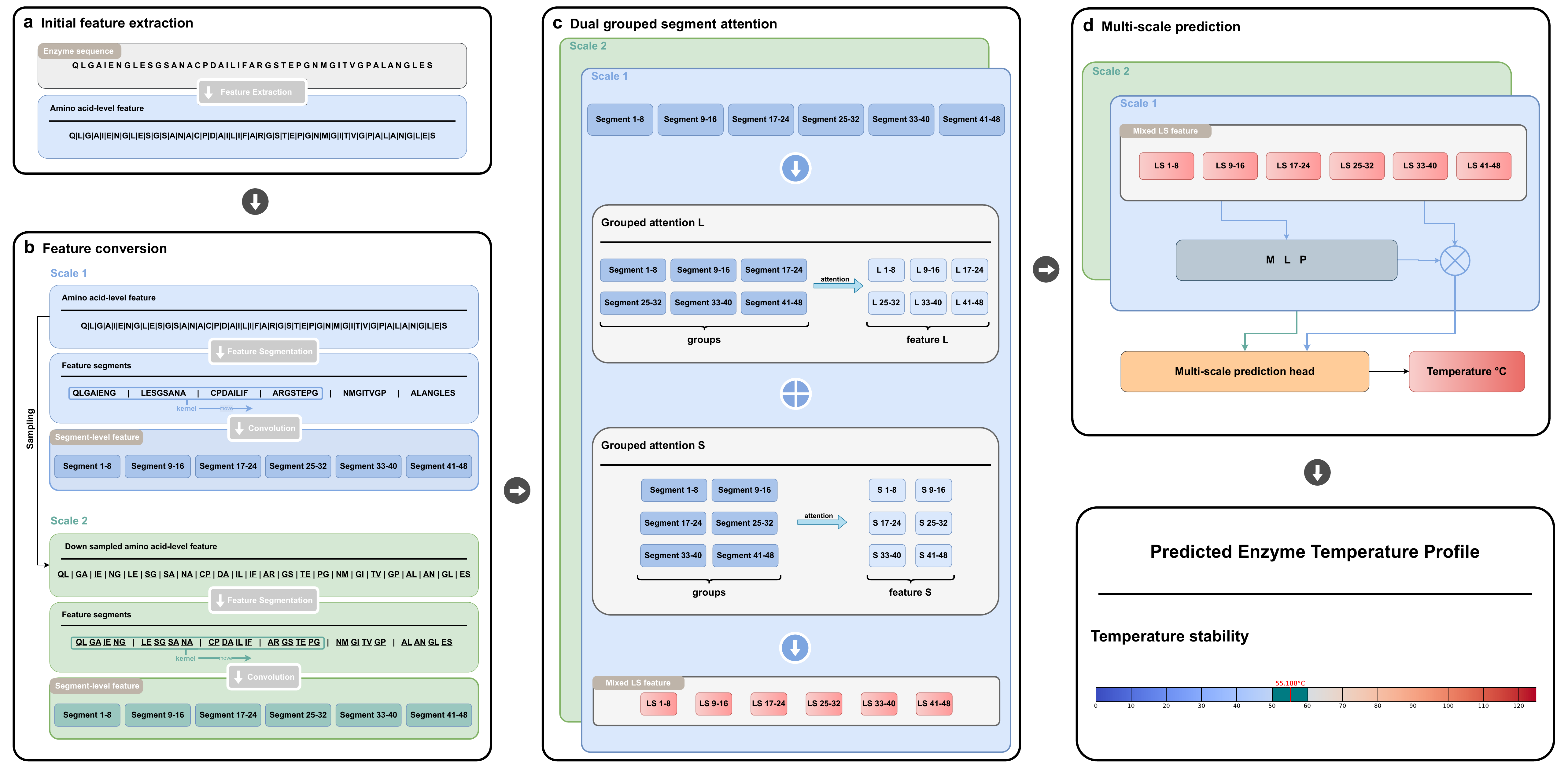}
\caption{\textbf{Detailed architecture of the segment transformer} \textbf{a} Feature extraction is performed to obtain initial amino acid-level features. \textbf{b} Conversion of amino acid-level features into segment-level features through sampling, segmentation, and convolution. \textbf{c} followed by grouped attention (L and S) and merging to produce a mixed LS feature. \textbf{d} Prediction of temperature stability values based on the mixed LS feature from two scales.}
\label{fig:framework}
\end{figure}

The segment transformer framework consists of two core modules: the feature conversion module (Fig.~\ref{fig:framework}b) and the dual grouped segment attention module (Fig.~\ref{fig:framework}c). First, the pretrained protein language model ESM-2 encodes the enzyme sequence into an initial feature of shape $L \times d$, where $L$ is the sequence length and $d$ is the embedding dimension \cite{lin2023evolutionary}. This initial feature is referred to as the amino acid-level feature, as each amino acid is mapped to a single embedding token.

Feature conversion aims to obtain multi-scale segment-level features. It performs sampling along the sequence-length dimension to generate a down-sampled amino acid-level feature of shape $L/2 \times d$. The initial feature and the down-sampled feature form scales 1 and 2, respectively. Following segmentation into fixed-length feature segments and convolution, both scales are converted into segment-level features. 

Subsequently, The segment-level features from both scales are passed into a stack of Dual Grouped Segment Attention (DGSA) blocks. Within each DGSA block, segments are reorganized into predefined groups, and grouped attention is applied independently to each group. Grouped attention L and S are used to capture different aspects of the grouped interactions. The resulting features from the L and S branches are summed to form a mixed LS feature, which captures diverse relationships between segments.

Finally, the multi-scale prediction module passes the mixed LS feature to a multilayer perceptron (MLP) to produce attention weights. These mixed LS features are multiplied by the attention weights to produce the prediction feature. The multi-scale prediction head then predicts the temperature stability value based on the prediction feature from both scales.

Furthermore, functional features have been integrated into the framework to enhance the practical value of segment transformer in enzyme engineering. As shown in the last part of Fig.~\ref{fig:framework}, segment transformer provides visualized prediction results for the prediction task. In addition to predicting specific temperature stability values, these results also provide a possible range of fluctation (indicated by the deep teal region), which serves as a valuable reference for researchers.

\subsection{Segment transformer}

The Segment Transformer is designed to leverage segment-level features, enabling the model to hierarchically capture both long- and short-range patterns within enzyme sequences. It operates through three primary stages: (1) feature conversion (Fig.~\ref{fig:framework}b), (2) dual grouped segment attention (Fig.~\ref{fig:framework}c), and (3) multi-scale prediction (Fig.~\ref{fig:framework}d). The details of each stage are described below.

\subsubsection{Feature Conversion}\label{sec_segment_feature}

To enable the model to learn patterns across varying levels of granularity, we convert amino acid-level features into segment-level representations. This process consists of three steps: sampling, sequence segmentation, and segment-wise convolution.

\begin{enumerate}
    \item 
    \textbf{Sampling:} Given an amino acid-level feature tensor $X \in \mathbb{R}^{B \times L \times D}$—where $B$ is the batch size, $L$ is the sequence length, and $D$ is the embedding dimension—we apply a series of strided 1D convolutional layers to obtain coarse-grained representations. Prior to convolution, the feature tensor is transposed to $X^\top \in \mathbb{R}^{B \times D \times L}$. Each convolutional layer $C_i$ (with kernel size 2 and stride 2) generates a downsampled representation:

    \begin{equation}
    X_i = C_i(X_{i-1}), \quad X_0 = X^\top, \quad i = 1, 2, \dots, S.
    \end{equation}

    where $S$ denotes the number of scales. The output is a set of multi-resolution feature maps: $\{X_0, X_1, \dots, X_S\}$.

    \item 
    \textbf{Segmentation:} Each feature map $X_i \in \mathbb{R}^{B \times D \times L_i}$ is transposed back to the shape $\mathbb{R}^{B \times L_i \times D}$ and reshaped into segments of fixed length $l_i$ (a predefined hyperparameter for scale $i$), yielding:

    \begin{equation}
    \tilde{X}_i \in \mathbb{R}^{B \times N_i \times l_i \times D}, \quad \text{where } N_i = \left\lfloor \frac{L_i}{l_i} \right\rfloor.
    \end{equation}

    This transforms the sequential features into a structured format where each segment captures localized context across $l_i$ residues.

    \item
    \textbf{Segment-wise Convolution:} To extract interactions across and within segments, we apply 2D convolutional layers to each segmented tensor $\tilde{X}_i$. A kernel of size $(k, l_i)$ is used along the segment dimension:

    \begin{equation}
    Y_i = \text{Conv2D}_i(\tilde{X}_i) \in \mathbb{R}^{B \times N_i \times D}.
    \end{equation}

    Here, the hyperparameter $k$ determines how many neighboring segments are considered during inter-segment interaction computation. This step yields segment-level representations that encode both intra- and inter-segment information. The final output is a collection of segment-level feature tensors $\{Y_0, Y_1, \dots, Y_S\}$, each corresponding to a different resolution scale.

\end{enumerate}

\subsubsection{Dual Grouped Segment Attention}\label{sec_segment_attention}

After converting amino acid-level inputs into multi-scale segment-level features, we apply a specialized attention mechanism—Dual Grouped Segment Attention—to capture both long- and short-range dependencies across segments.

Given a segment-level feature tensor $Y_i \in \mathbb{R}^{B \times N_i \times D}$, where $B$ is the batch size, $N_i$ is the number of segments at scale $i$, and $D$ is the feature dimension, we reshape the input into a 4D tensor structured as groups of segments with a fixed range $G_i^S$ (a predefined hyperparameter):

\begin{equation}
Y_i' \in \mathbb{R}^{B \times G_i^L \times G_i^S \times D}, \quad \text{where } G_i^L = \left\lfloor \frac{N_i}{G_i^S} \right\rfloor.
\end{equation}

This structure enables attention to be performed from two complementary perspectives:

\begin{itemize}
    \item 
    \textbf{Grouped Attention S:} Flattening the batch and long-range group dimensions results in a tensor of shape $(B \cdot G_i^L, G_i^S, D)$. This format emphasizes local context within each short-range group. Short-range attention is then applied to the reshaped tensor:

    \begin{equation}
    Z_i^S = \text{Attention}_{\text{short}}(Y_i^{(S)}) \in \mathbb{R}^{B \cdot G_i^L \times G_i^S \times D}.
    \end{equation}

    \item 
    \textbf{Grouped Attention L:} Reshaping the 4D tensor to a format of $(B \cdot G_i^S, G_i^L, D)$ enables the model to learn broader contextual relationships among segments of the sequence:

    \begin{equation}
    Z_i^L = \text{Attention}_{\text{long}}(Y_i^{(L)}) \in \mathbb{R}^{B \cdot G_i^S \times G_i^L \times D}.
    \end{equation}
    
\end{itemize}

After obtaining the attention-enhanced features, we reshape both $Z_i^S$ and $Z_i^L$ back to their original 4D grid shape and sum them to integrate local and broad contextual information. The resulting tensor is then flattened back into a sequence format:

\begin{equation}
Z_i = Z_i^L + Z_i^S \in \mathbb{R}^{B \times G_i^L \times G_i^S \times D}.
\end{equation}

\begin{equation}
\hat{Y}_i = \text{Flatten}(Z_i) \in \mathbb{R}^{B \times N_i \times D}.
\end{equation}

The Dual Grouped Segment Attention is applied independently to each scale. Given a list of segment-level features $\{Y_0, Y_1, \dots, Y_S\}$, the attention module is applied at each resolution:

\begin{equation}
\{\hat{Y}_0, \hat{Y}_1, \dots, \hat{Y}_S\} = \{\text{Attention}_0(Y_0), \text{Attention}_1(Y_1), \dots, \text{Attention}_S(Y_S)\}.
\end{equation}

These refined features encode enhanced contextual information across multiple scales and are used in subsequent prediction modules.

\subsubsection{Multi-Scale Prediction}\label{sec_multi_scale_prediction}

Following Dual Grouped Segment Attention, the model produces multiple sets of refined segment-level features across different scales. To integrate these multi-scale features into a unified representation for final prediction, we employ attention-based pooling followed by a multi-scale prediction head.

For each scale $i$, given the attended segment-level features $\hat{Y}_i \in \mathbb{R}^{B \times N_i \times D}$, we compute attention weights to perform weighted aggregation of segment features. These weights are derived using a two-layer feedforward network with a non-linear activation. The pooled feature vector at scale $i$ is computed as follows:

\begin{align}
A_i &= \tanh(W_1 \hat{Y}_i). \\
\alpha_i &= \text{softmax}(W_2 A_i). \\
z_i &= \sum_{j=1}^{N_i} \alpha_{i,j} \hat{Y}_{i,j}.
\end{align}

Here, $W_1$ and $W_2$ are learnable parameters, and $\alpha_{i,j}$ denotes the attention weight assigned to the $j$-th segment at scale $i$. This attention pooling mechanism enables the model to selectively emphasize informative segments based on their relevance.

After pooling, we obtain a set of scale-specific representations $\{z_0, z_1, \dots, z_S\}$, each capturing contextual information at a different resolution. These representations are independently passed through scale-specific prediction layers. The final prediction is then computed by aggregating the outputs across all scales:

\begin{align}
\hat{y}_i &= W_{\text{reg},i} z_i + b_{\text{reg},i}. \\
\bar{\alpha} &= \frac{1}{S+1} \sum_{i=0}^{S} \alpha_i. \\
\hat{y} &= \frac{1}{S+1} \sum_{i=0}^{S} \hat{y}_i. \\
\hat{y}_{\text{max}} &= \max(\hat{y}_0, \hat{y}_1, \dots, \hat{y}_S). \\
\hat{y}_{\text{min}} &= \min(\hat{y}_0, \hat{y}_1, \dots, \hat{y}_S).
\end{align}

This multi-scale fusion strategy leverages complementary information from various segment resolutions, enhancing the robustness and accuracy of the final prediction.

Additionally, the attention weights $\{\alpha_i\}$ reflect the relative importance of individual segments at each scale. To derive a unified segment importance profile, we average the attention weights across all scales. The resulting vector $\bar{\alpha}$ provides interpretability by highlighting the segments that contribute most significantly to the prediction, offering insights into biologically or structurally important regions.

\subsection{Experimental Setup and Loss Functions}\label{sec_setup&lossfunc}

All experiments were performed on a server running Ubuntu 22.04.5 LTS, equipped with Python 3.12.7 and PyTorch 2.5.1. The hardware environment included an NVIDIA H100 GPU with CUDA version 12.2. The Segment Transformer model was implemented in PyTorch and optimized using the AdamW optimizer.

\subsubsection{Weighted RMSE Loss for Temperature Stability Prediction}

To address the imbalanced distribution of temperature stability values, we adopted a weighted root mean square error (Weighted RMSE) loss. This formulation allows the model to emphasize underrepresented temperature intervals during training, which is particularly important in the context of enzyme stability prediction. The Weighted RMSE is defined as:

\begin{equation}
\text{WeightedRMSE} = \sqrt{ \frac{1}{N} \sum_{i=1}^{N} w(t_i) \cdot (\hat{y}_i - y_i)^2 }. 
\end{equation}

Here, $\hat{y}_i$ and $y_i$ denote the predicted and true scalar temperatures for the $i$-th sample. The temperature range is partitioned into $K$ intervals $\mathcal{R}_k = [a_k, b_k)$, each associated with a weight $w_k$ for $k = 1, 2, \dots, K$. The weight for sample $i$ is given by $w(t_i) = w_k$ if $t_i \in \mathcal{R}_k$.

\subsection{Evaluation Metrics}\label{subsec_metrics}

To access model performance, we used four evaluation metrics to assess both error magnitude and correlation: Pearson correlation coefficient, Spearman correlation coefficient, mean absolute error (MAE), and root mean square error (RMSE). Given $N$ samples with predicted values $\hat{y}_i$ and ground truth values $y_i$, the metrics are defined as follows:

\begin{equation}
\text{Pearson} = \frac{\sum_{i=1}^{N} (\hat{y}_i - \bar{\hat{y}})(y_i - \bar{y})}{\sqrt{\sum_{i=1}^{N} (\hat{y}_i - \bar{\hat{y}})^2} \sqrt{\sum_{i=1}^{N} (y_i - \bar{y})^2}}.
\end{equation}

\begin{equation}
\text{Spearman} = \text{Pearson}(\text{rank}(\hat{y}_i), \text{rank}(y_i)).
\end{equation}

\begin{equation}
\text{MAE} = \frac{1}{N} \sum_{i=1}^{N} \left| \hat{y}_i - y_i \right|.
\end{equation}

\begin{equation}
\text{RMSE} = \sqrt{ \frac{1}{N} \sum_{i=1}^{N} (\hat{y}_i - y_i)^2 }.
\end{equation}

These metrics collectively evaluate the accuracy and consistency of the scalar temperature predictions.

\subsection{t-SNE Visualization}\label{sec_tsne}

To gain insight into the internal representations learned by segment transformer, we utilized the \textit{t-distributed stochastic neighbor embedding} (t-SNE) algorithm for visualization \cite{JMLR:v9:vandermaaten08a}. t-SNE is a nonlinear dimensionality reduction technique that projects high-dimensional data into a two-dimensional space, preserving local neighborhood structures to facilitate intuitive visual interpretation. Three types of intermediate embeddings were extracted from the trained segment transformer model on the test set:

\begin{enumerate}
    \item Initial segment-level features produced by the segment-wise convolutional layers.
    \item Contextualized representations from the segment transformer.
    \item Final pooled feature vectors used for prediction.
\end{enumerate}

Each of these embeddings was projected into two dimensions using t-SNE with a perplexity of 30. Data points in the resulting plots were colored according to their associated temperature labels, enabling visual assessment of feature separation with respect to thermal properties. The t-SNE visualizations reveal a progressive refinement of the feature space throughout the network. While the convolutional features exhibit coarse clustering by temperature, the representations become increasingly structured and temperature-discriminative in the transformer and final pooled embeddings. These visualizations provide qualitative evidence that segment transformer effectively captures and organizes thermophilic patterns in enzyme sequences.

\subsection{Application Study of Segment Transformer}\label{sec_experiment}
In this study, a cutinase (UniProt ID: A0A075B5G4) from \textit{Humicola insolens} (HiC) was selected as the case enzyme. A thermostability enhancement strategy was designed based on predictions of segment transformer and subsequently validated through biological experiments.

\subsubsection{Strains and Reagents}

The recombinant plasmid pET20b(+)-\textit{hic} was used as the backbone vector for cloning and heterologous expression. \textit{Escherichia coli} JM109 (Novagen, Madison, WI, USA) and BL21(DE3) pLysS (TransGen Biotech Co., Ltd., Beijing, China) were employed for library construction and protein expression, respectively. Strains were cultivated at 37°C in LB medium (1\% w/v tryptone, 0.5\% w/v yeast extract, and 1\% w/v NaCl), while protein production was carried out at 25°C in TB medium (2.4\% w/v tryptone, 1.2\% w/v yeast extract, 0.5\% w/v glycerol, 0.231\% w/v KH\textsubscript{2}PO\textsubscript{4}, and 1.643\% w/v K\textsubscript{2}HPO\textsubscript{4}). Plasmid extraction kits were purchased from TianGen Biotech (Beijing, China). The substrate 4-nitrophenyl butyrate (pNPB) was obtained from Aladdin (Shanghai, China), and all other analytical-grade reagents were sourced from Sinopharm Chemical Reagent Co., Ltd. (Beijing, China).

\subsubsection{Construction, Expression, and Purification}

Single-point mutants were constructed using the pET20b(+)-\textit{hic} plasmid as a template via high-fidelity PCR. Reactions were performed using Phanta Max Super-Fidelity DNA Polymerase (2× Phanta Master Mix, Dye Plus; Vazyme Biotech Co., Ltd., Nanjing, China) and mutagenic primers (Supplementary Table 1). PCR products were digested with DpnI to eliminate template DNA and transformed into \textit{E. coli} JM109 competent cells. Positive clones were selected on LB agar plates containing ampicillin (100~\textmu{}g/mL) and verified by Sanger sequencing. Confirmed plasmids were then transformed into \textit{E. coli} BL21(DE3) pLysS cells for protein expression.

Transformed colonies were grown overnight in 10 mL LB medium at 37°C (220 rpm), followed by inoculation (5\% v/v) into 50 mL TB medium. Cultures were incubated at 37°C (220 rpm) for 2 h, and protein expression was induced by reducing the temperature to 25°C for 48 h under the same agitation conditions.

\subsubsection{Determination of Enzymatic Thermal Stability}

The thermostability of wild-type and mutant HiC enzymes was assessed by measuring residual activity after heat treatment. Enzymes were diluted in citrate buffer (1000 mM, pH 7.0) and incubated at 60°C for 5 minutes. Residual activity was measured using 4-nitrophenyl butyrate (pNPB) as the substrate. The assay followed the spectrophotometric method described by Chen et al. (2008), with one unit of activity defined as the amount of enzyme that releases 1~\textmu{}mol of 4-nitrophenol per minute under assay conditions. All experiments were performed in triplicate using independent biological replicates.

\subsubsection{Homology Modeling and Molecular Dynamics (MD) Simulations}

Homology models of cutinase mutants were constructed using SWISS-MODEL, with the crystallographic structure of HiC (PDB ID: 4OYY, 3.0 Å resolution) as the template. Molecular dynamics (MD) simulations were conducted at 330 K to assess enzyme flexibility, following an adapted protocol from Rao et al.~\cite{rao2022trehalose}. Protein models were immersed in a TIP3P water environment, and simulations were performed using the Amber ff14SB force field. Systems were neutralized with sodium and chloride ions prior to simulation. Energy minimization involved two phases: 5,000 steepest descent iterations followed by 5,000 conjugate gradient cycles. The system was gradually heated over 50 ps, followed by 200 ps of NPT equilibration to stabilize pressure and density. Production runs spanned 100 ns to capture molecular trajectories, using a 2 fs timestep. Covalent bonds were constrained via the SHAKE algorithm, while temperature and pressure were regulated using Langevin and Berendsen methods, respectively. Subsequently, RMSD, RMSF, Rg, and SASA were quantified using the cpptraj module~\cite{roe2013ptraj}.

\section{Results and Discussion}

\subsection{Architecture Optimization}\label{res_segment_feature}

\begin{figure}
    \centering
    \includegraphics[width=\linewidth]{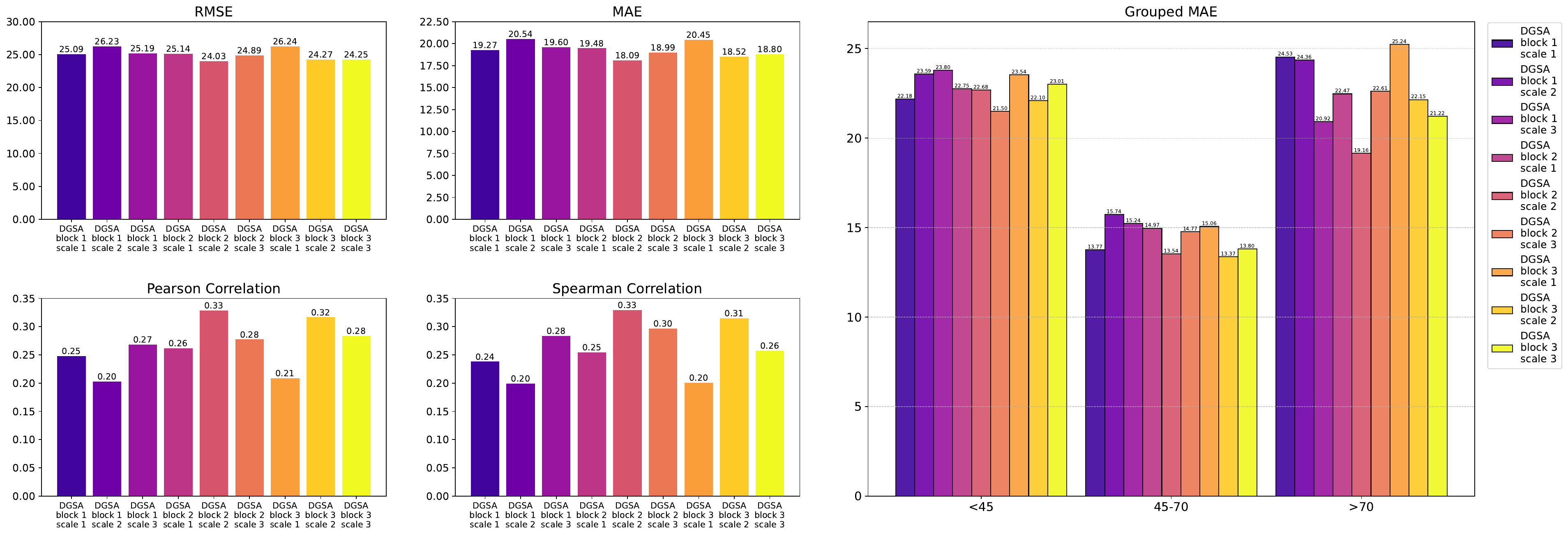}
    \caption{\textbf{Optimization of the Segment Transformer architecture.} Root Mean Square Error (RMSE), Mean Absolute Error (MAE), Pearson correlation, Spearman correlation, and Grouped MAE for different model configurations on the temperature stability test set. Temperature groupings are based on the data distribution: $<$45°C, 45--70°C, and $\geq$70°C.}
    \label{fig:ablation}
\end{figure}

To identify the most effective architecture for enzyme temperature stability prediction, we conducted a comprehensive set of ablation experiments. A key insight from these experiments is that transforming amino acid-level features into segment-level features significantly enhances predictive performance. Building on this foundation, we incorporated multi-scale features and introduced the Dual Grouped Segment Attention (DGSA) block to better exploit hierarchical sequence-level information.

To evaluate model performance, we used four standard regression metrics: Root Mean Square Error (RMSE), Mean Absolute Error (MAE), Pearson correlation, and Spearman correlation. Given the imbalanced distribution of temperature stability values in the dataset, we additionally defined the Grouped MAE metric to evaluate model robustness across different temperature ranges. This metric assesses performance within low ($<45^\circ$C), moderate (45--70$^\circ$C), and high ($\geq$70$^\circ$C) thermal regimes, which are critical for enzyme engineering applications.

Based on a systematic comparison across different architectural configurations (varying the number of feature scales and DGSA blocks), we selected a final design incorporating two feature scales and two DGSA blocks (Fig.~\ref{fig:ablation}). This configuration, referred to as the \textit{Segment Transformer}, achieved the best overall performance with an RMSE of 24.03, MAE of 18.09, and both Pearson and Spearman correlations of 0.33.

Notably, the architecture also achieved the lowest Grouped MAE across all temperature intervals: 21.50 in the low range ($<45^\circ$C), 13.54 in the moderate range (45--70$^\circ$C), and 19.16 in the high range ($\geq$70$^\circ$C), indicating strong generalization across the full temperature spectrum.

\subsection{Performance Evaluation}

We compare the Segment Transformer with five classic deep learning models and three state-of-the-art enzyme temperature stability predictors: BiLSTM, CNN, Transformer, Light Attention, RNN, TemStaPro, Seq2Topt, and DeepET \cite{schuster1997bidirectional, collobert2011natural, vaswani2017attention, stark2021light, elman1990finding, pudvziuvelyte2024temstapro, qiu2025seq2topt, li2022learning}. To ensure a fair comparison, all models were trained using the same reweighted RMSE loss function (see Section~\ref{sec_setup&lossfunc}), which mitigates the effects of data imbalance by discouraging overfitting to the dominant mid-temperature range (45--70$^\circ$C) and encouraging learning from underrepresented temperature extremes.

Model performance is evaluated using Root Mean Square Error (RMSE), Mean Absolute Error (MAE), Pearson correlation, and Spearman correlation. To further assess model robustness across varying thermal regimes, we introduce the Grouped MAE metric, which evaluates prediction accuracy within stratified temperature intervals. Additionally, we employ scatter plots and t-SNE visualizations to provide qualitative insights into the learned representations and predictive patterns.

\begin{figure}[htbp]
    \centering
    \includegraphics[width=0.45\linewidth]{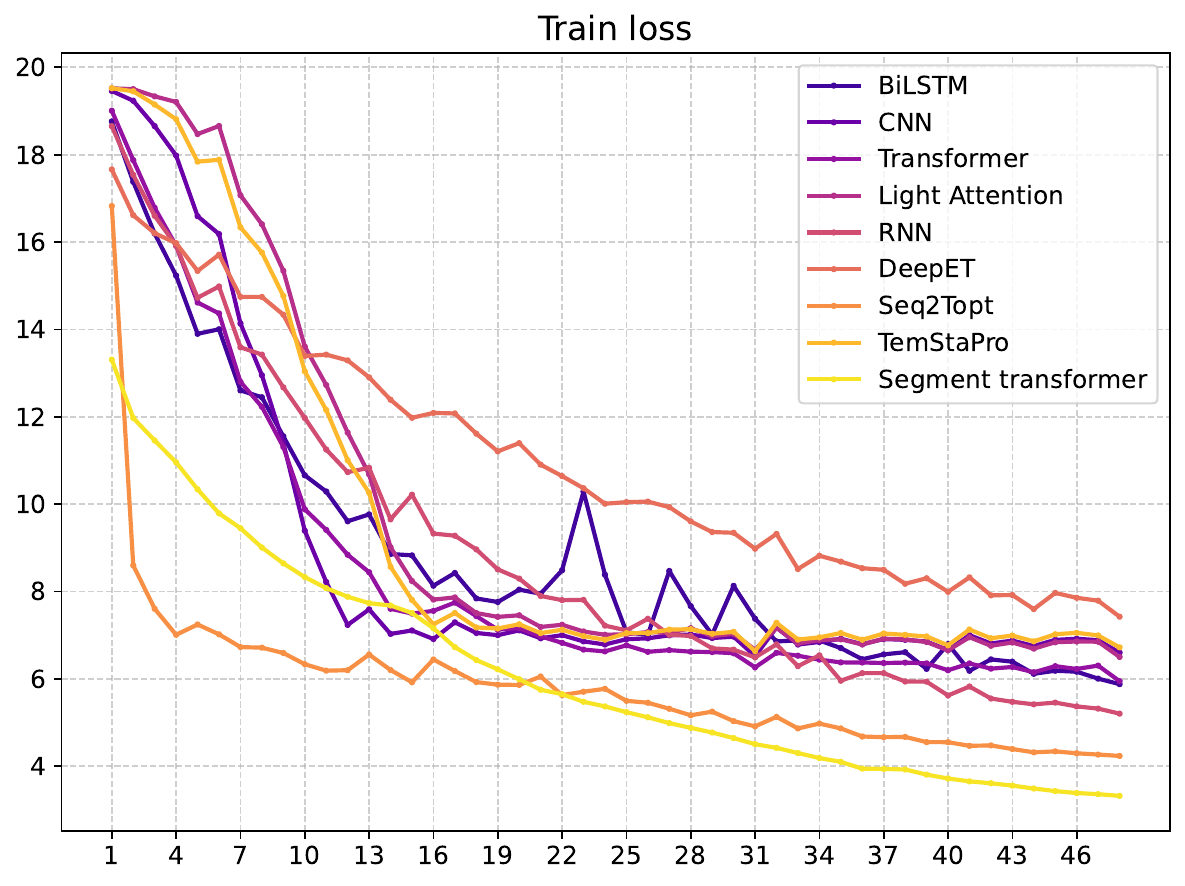}
    \includegraphics[width=0.45\linewidth]{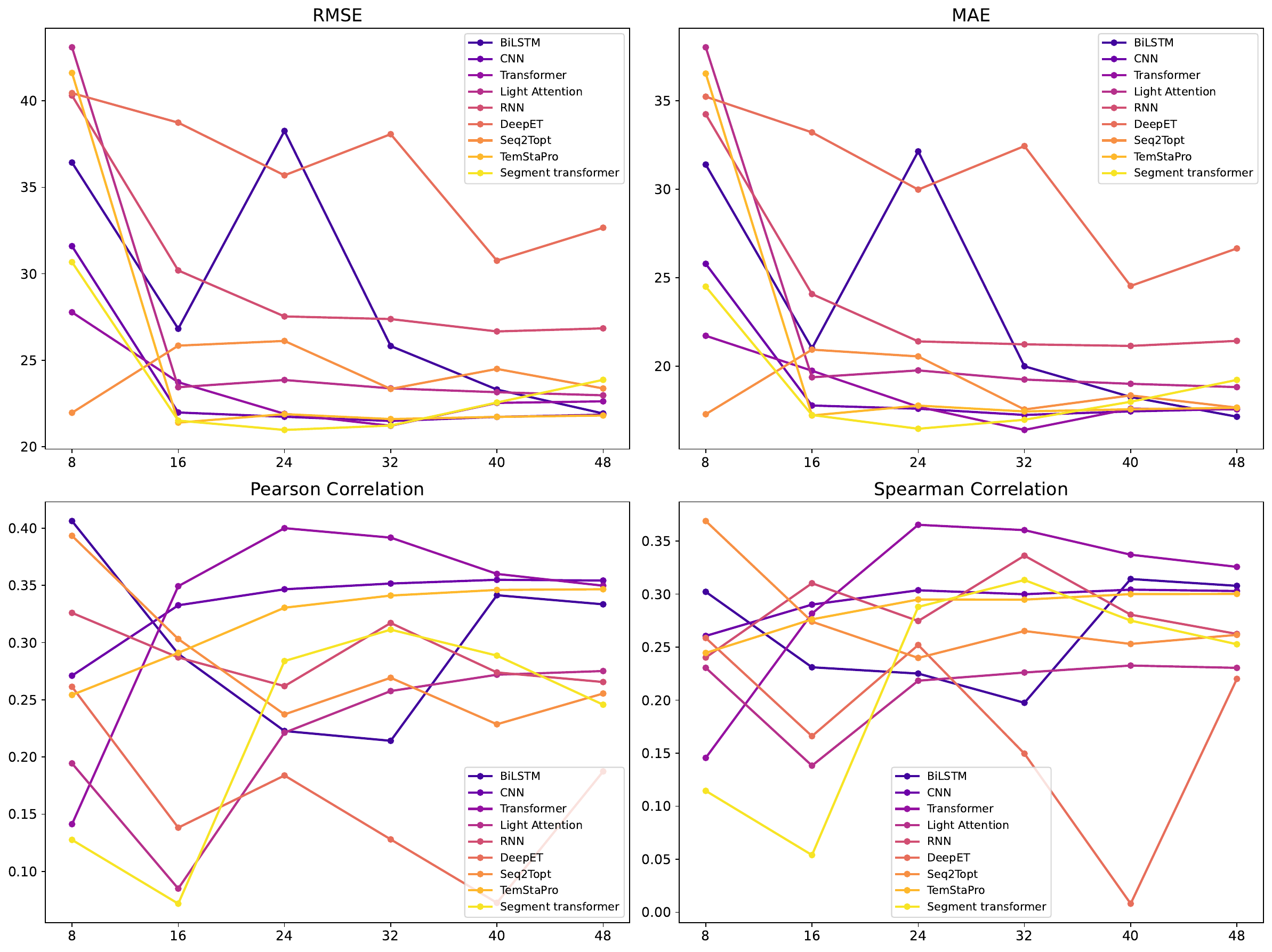}
    \caption{\textbf{Training dynamics and validation performance.} RMSE, MAE, Pearson correlation, and Spearman correlation were evaluated every 8 epochs on the validation set.}
    \label{fig:val}
\end{figure}

\begin{figure}[htbp]
    \centering
    \includegraphics[width=1\linewidth]{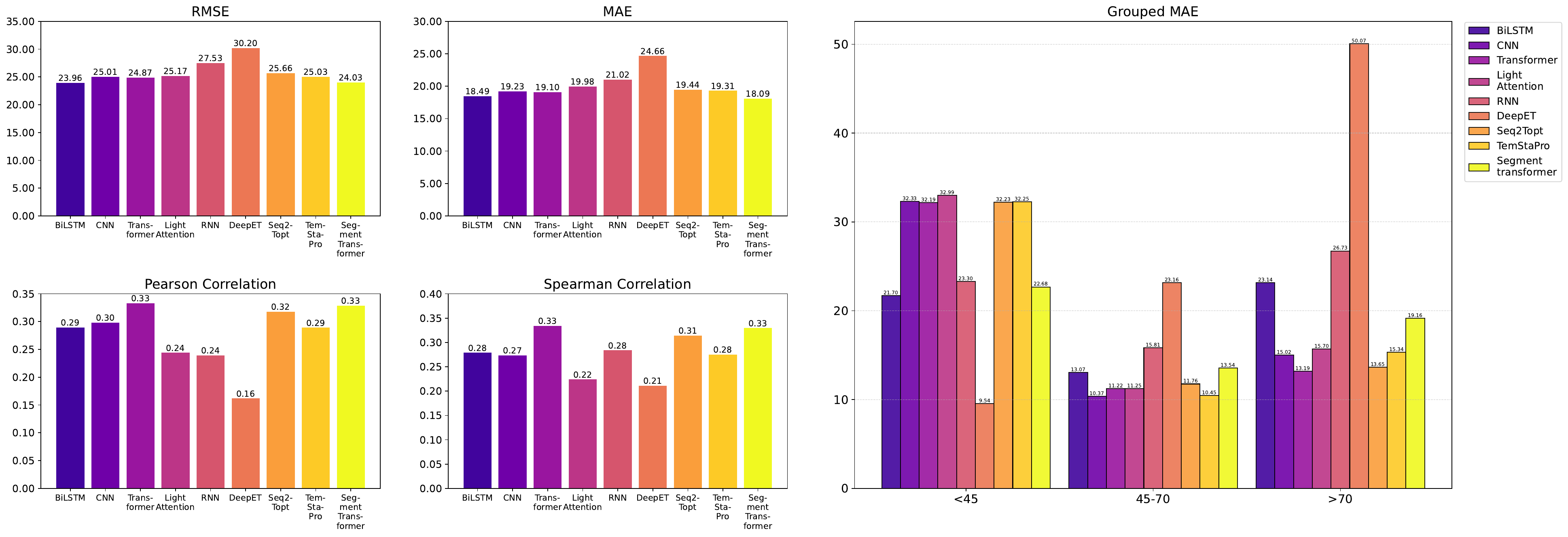}
    \caption{\textbf{Performance on the temperature stability test set.} Evaluation of RMSE, MAE, Pearson correlation, Spearman correlation, and Grouped MAE across all models.}
    \label{fig:test}
\end{figure}

As shown in Fig.~\ref{fig:val}, the training loss of the Segment Transformer steadily decreases over time, and validation metrics are recorded every 8 epochs. The best validation results achieved by the Segment Transformer are RMSE = 20.26, MAE = 15.75, Pearson correlation = 0.319, and Spearman correlation = 0.314. The corresponding checkpoint was selected for final testing.

On the test set (Fig.~\ref{fig:test}), the Segment Transformer outperforms all other models, achieving an RMSE of 24.03, MAE of 18.09, and both Pearson and Spearman correlations of 0.33. Although BiLSTM also performs competitively, relying solely on overall metrics may mask performance variability across different temperature ranges. Grouped MAE offers a more detailed view of model behavior in low (<45°C), medium (45–70°C), and high ($\geq70$°C) regimes.

\begin{figure}[htbp]
    \centering
    \includegraphics[width=1\linewidth]{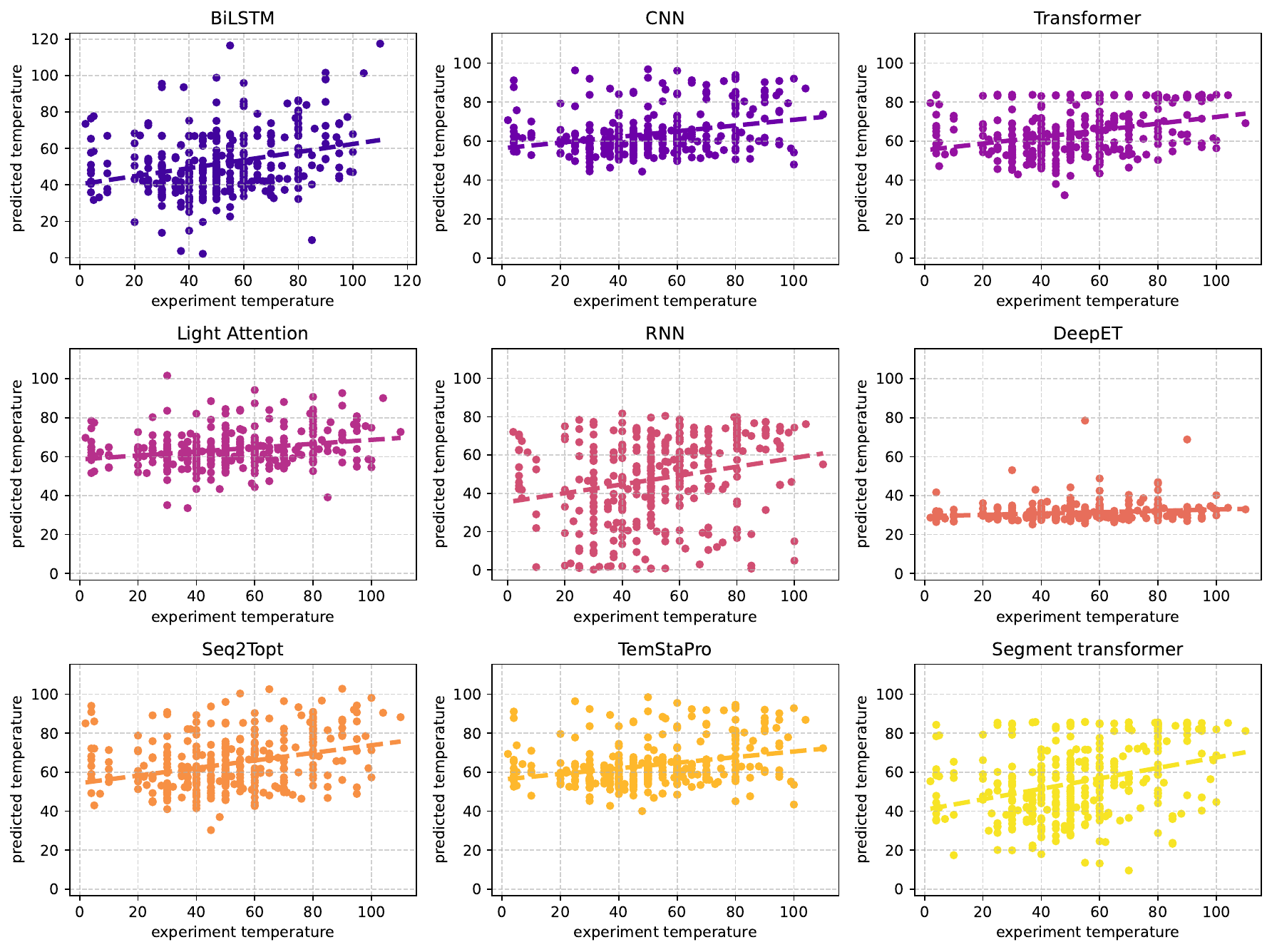}
    \caption{\textbf{Scatter plots of predicted vs. actual values} for the Segment Transformer and comparison models.}
    \label{fig:scatter}
\end{figure}

The scatter plots in Fig.~\ref{fig:scatter} further illustrate performance disparities. CNN, Transformer, Light Attention, Seq2Topt, and TemStaPro exhibit limited prediction ranges, struggling to predict low-temperature stability. Conversely, RNN favors low-temperature predictions but underperforms at high temperatures. DeepET shows an even narrower range, with predictions clustering between 20–40°C—likely due to its reliance on handcrafted features and the lack of pretrained embeddings. Overall, BiLSTM and Segment Transformer demonstrate the most balanced predictive behavior, as evidenced by their broader and more accurate prediction spans.

While BiLSTM performs well overall, the Segment Transformer consistently yields higher correlation coefficients and exhibits superior Grouped MAE in the high-temperature regime ($\geq70$°C). Specifically, both models have comparable MAEs in the <45°C and 45–70°C intervals, but the Segment Transformer achieves a significantly lower MAE in the high-temperature range. These results highlight the Segment Transformer’s capacity to maintain stable and accurate predictions across all thermal categories, which is crucial for general-purpose enzyme engineering.

\begin{figure}[htbp]
    \centering
    \includegraphics[width=1\linewidth]{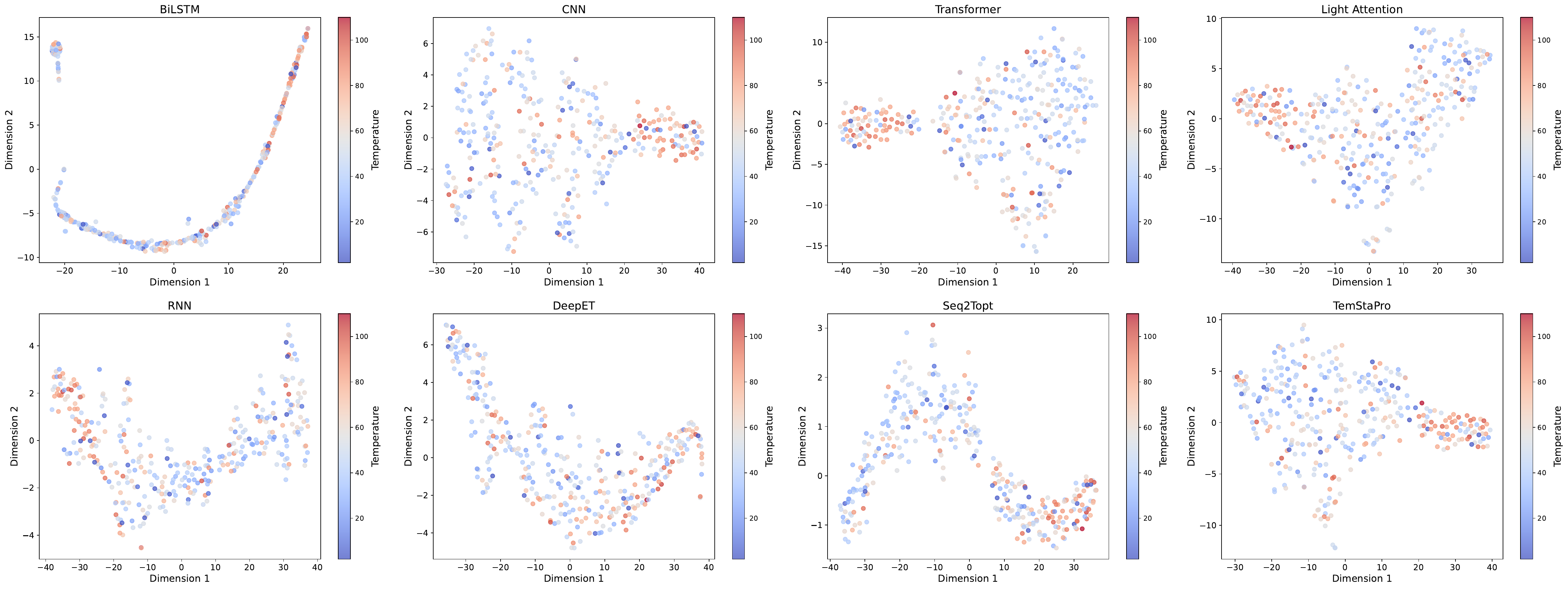}
    \includegraphics[width=0.3\linewidth]{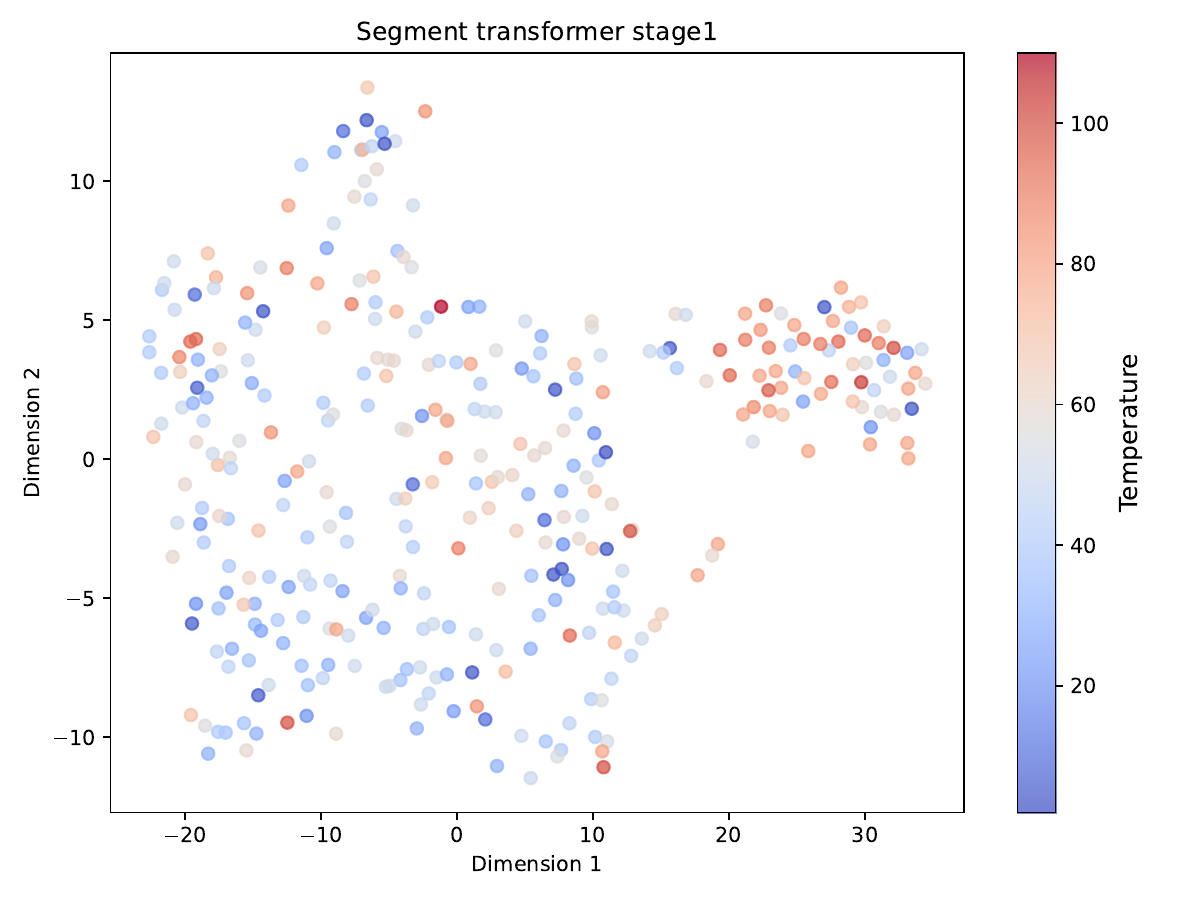}
    \includegraphics[width=0.3\linewidth]{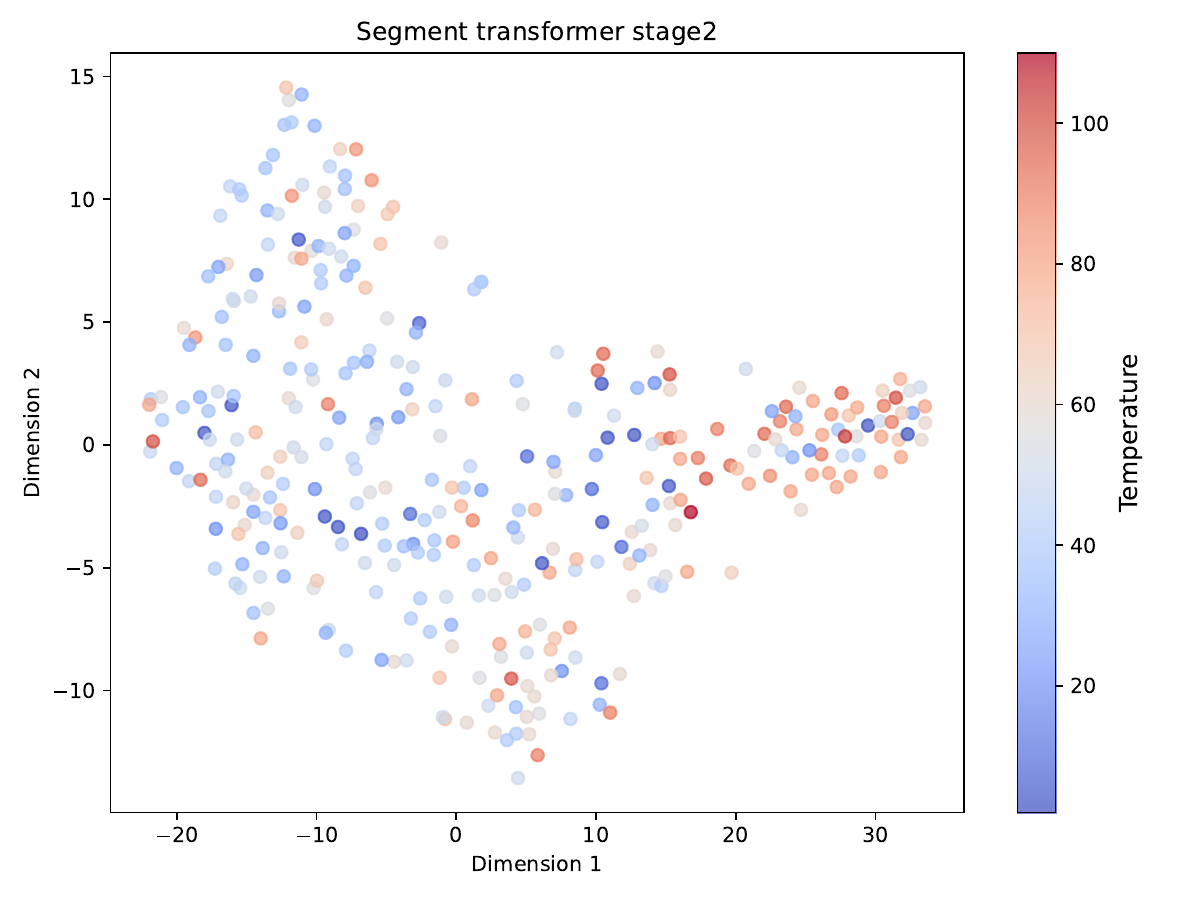}
    \includegraphics[width=0.3\linewidth]{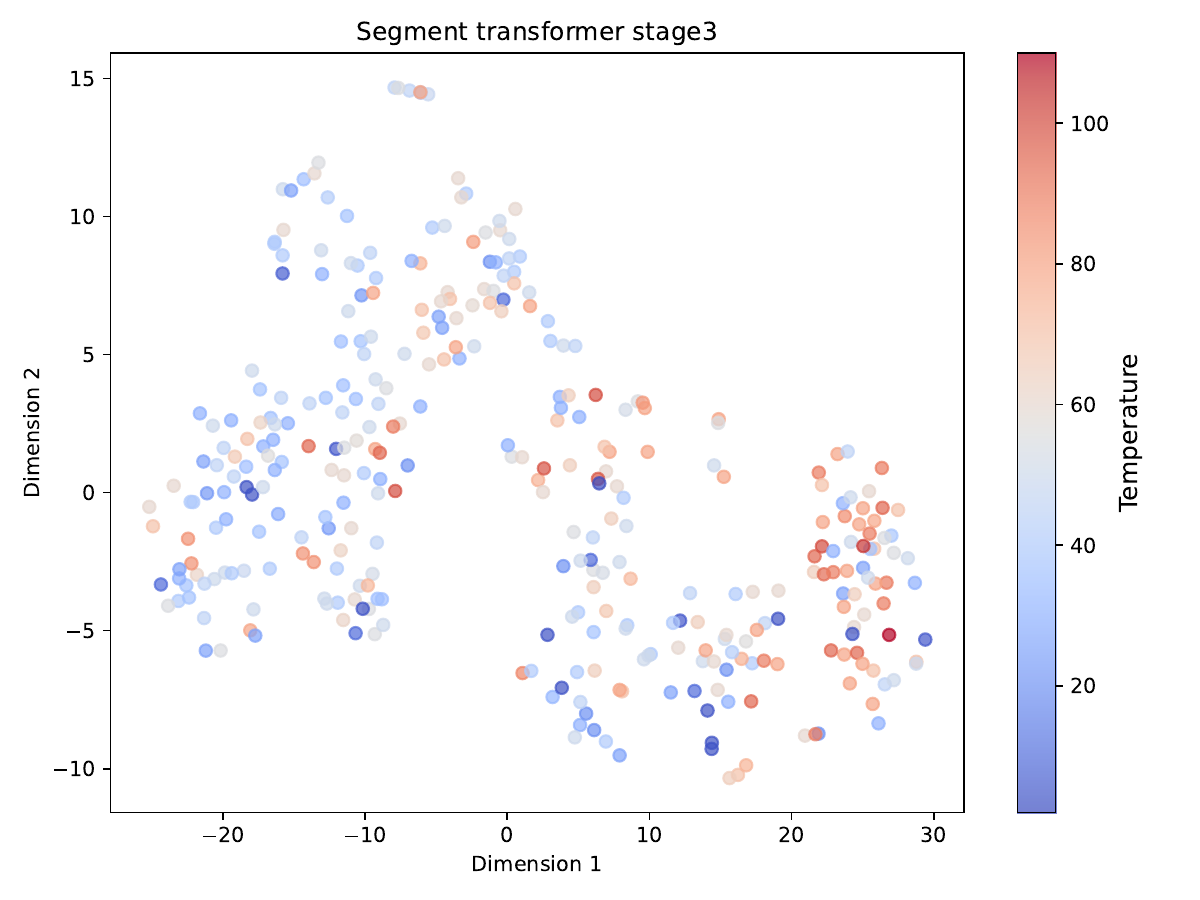}
    \caption{\textbf{t-SNE visualizations.} Top: t-SNE plots of learned features from comparison models. Bottom: visualization of feature representations across three stages in the Segment Transformer.}
    \label{fig:tsne}
\end{figure}

To better understand the representation learning of each model, we applied t-SNE to visualize the learned feature spaces (Fig.~\ref{fig:tsne}). In comparison models, high-stability (red) and low-stability (blue) samples often overlap, indicating poor separation in the latent space. In contrast, the Segment Transformer shows progressive improvement in clustering through its inference pipeline. The last three subfigures demonstrate the feature refinement stages in the Segment Transformer, showing t-SNE plots of: (1) the initial segment features, (2) DGSA-transformed features, and (3) final output features. The enhanced separation in the final stage confirms the model's effective hierarchical learning and generalization.

\subsection{Applications of segment transformer in Thermal Property-Related Enzyme Engineering}\label{sec_evolution}

\begin{figure}[htbp]
\centering
\includegraphics[width=\linewidth]{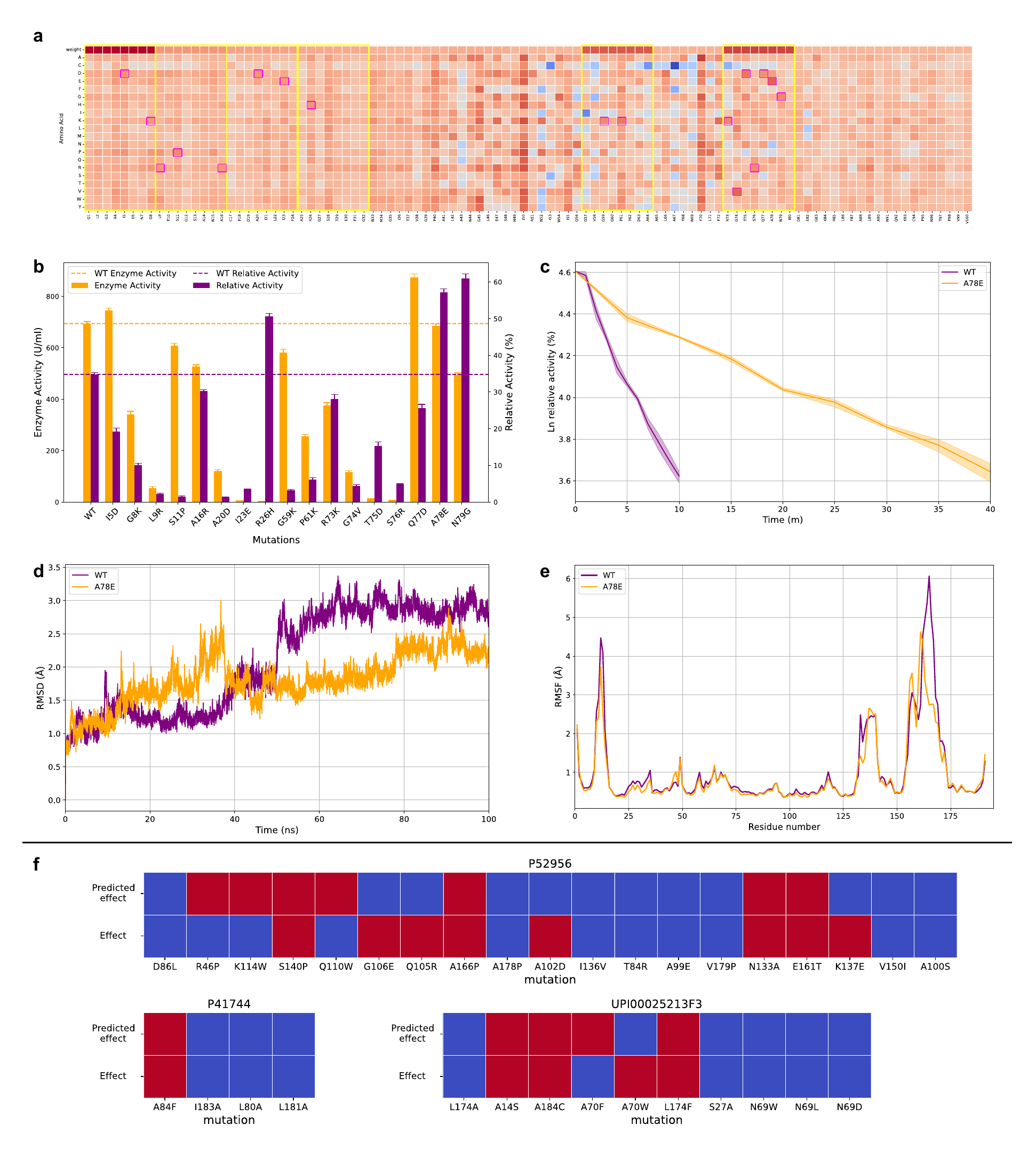}
\caption{\textbf{Segment transformer Guided Thermostability Enhancement Strategy and Experimental Results.} \textbf{a} Importance scores and temperature scores predicted by segment transformer. \textbf{b} Enzyme activity and relative activity of all mutations and the wild type (WT). \textbf{c} Determination of the half-life of WT and A78E at 60°C. \textbf{d} Root mean square deviation (RMSD) of WT and A78E. \textbf{e} Root mean square fluctuation (RMSF) of WT and A78E. \textbf{f} Predicted and actual effects of mutations on three additional case enzymes.}
\label{fig_application}
\end{figure}

In this study, a cutinase (UniProt ID: A0A075B5G4) derived from \textit{Humicola insolens} (HiC) was selected as the experimental case. This enzyme, belonging to the $\alpha/\beta$ hydrolase family, exhibits multifunctional catalytic capabilities, including hydrolysis of natural cutin into fatty acid monomers, efficient degradation of soluble esters, insoluble triglycerides, and synthetic polyesters, demonstrating broad substrate promiscuity. Owing to its versatile catalytic properties—encompassing insoluble lipid decomposition, polyester/amide bond cleavage, ester synthesis, and transesterification—the industrial applications of this cutinase were systematically explored. These applications span detergent additive development, food processing optimization, environmental remediation, textile bio-polishing, and pulp modification. However, the sustained high-temperature conditions inherent to industrial processes limit its broader application, underscoring the need to develop thermostability-enhanced variants to meet operational demands in energy-intensive biomanufacturing systems. Consequently, we employed segment transformer to design a mutation strategy to enhance the thermostability of this enzyme.

Specifically, segment transformer predicts importance scores and temperature scores to guide thermostability-related enzyme engineering. Importance scores, shown in the weight row of Fig.~\ref{fig_application}a, reflect the thermal property contribution of each segment. Temperature scores, shown in the remaining rows of Fig.~\ref{fig_application}a, indicate potential improvement in thermostability (positive scores, red) or a detrimental effect (negative scores, blue). Color intensity reflects the magnitude of the predicted mutation impact. Based on these importance and temperature scores, the following criteria were used to select mutations:

\textit{Candidate segments were prioritized based on a threshold of an overall segment importance score exceeding 20, yielding six qualified segments (marked by yellow frames in Fig.~\ref{fig_application}a). Within these segments, amino acid positions exhibiting the highest individual temperature scores ($>$50) were selected as candidate mutation sites (marked by pink frames in Fig.~\ref{fig_application}a).}

Eventually, seventeen candidate mutations were identified for potential thermostability enhancement based on segment transformer’s predictions. The experimental results are shown in Fig.~\ref{fig_application}b, where orange bars represent enzyme activity, purple bars represent relative enzyme activity after heat treatment, and error bars indicate the standard deviation from triplicate experiments. The results show that mutants I5D, S11P, A16R, G59K, Q77D, A78E, and N79G retained over 60\% of wild-type (WT) activity. Although R26H exhibited moderate improvement in thermostability, it was excluded from further analysis due to significantly reduced activity. In contrast, Q77D showed a slight decrease in thermostability compared to WT but achieved a 1.26-fold increase in activity. Among these, A78E and N79G showed significantly improved thermostability, with relative activity increased by 1.64-fold and 1.75-fold after heat treatment, respectively. Since A78E exhibited minimal reduction in enzyme activity, we selected this mutation for further analysis.

The capacity of enzymes to maintain catalytic activity following irreversible denaturation is termed kinetic stability. To systematically investigate the thermal inactivation kinetics of WT and A78E, temperature-dependent stability profiles were determined by measuring the thermal half-life ($t_{1/2}$) at 60°C. As shown in Fig.~\ref{fig_application}c, the WT demonstrated a $t_{1/2}$ of 6.8 minutes, whereas the A78E mutant exhibited a significantly prolonged $t_{1/2}$ of 29.5 minutes, representing a 3.9-fold enhancement in thermal stability compared to WT.

Molecular dynamics (MD) simulations of the WT enzyme and the A78E mutant were conducted at 330 K for 100 ns. The root mean square deviation (RMSD), a critical parameter for evaluating conformational stability, reflects changes in the unfolding enthalpy and entropy of proteins. Thermally stable enzymes typically exhibit lower RMSD values due to enhanced structural rigidity. As shown in Fig.~\ref{fig_application}d, both WT and A78E reached equilibrium after 50 ns, with the mutant displaying significantly reduced structural fluctuations, suggesting that the A78E mutation increased structural rigidity, a key contributor to its improved thermostability. Root mean square fluctuation (RMSF) analysis, which quantifies atomic positional deviations from average coordinates, further corroborated these findings. Lower RMSF values in A78E (Fig.~\ref{fig_application}e) indicated enhanced local stability, particularly in the previously hyperflexible 160--170 loop region. This loop, located near the catalytic triad, is critical for substrate binding; its stabilization likely minimized unproductive conformational shifts during catalysis. The combined reduction in RMSD and RMSF values demonstrated that the A78E mutation enhanced global structural rigidity while preserving functional flexibility at the active site. This balance is essential for industrial enzymes operating under thermal stress, as excessive rigidity can impair catalytic efficiency.

While these results collectively highlight segment transformer’s ability to guide thermostability-related enzyme engineering, experimental data from a single enzyme may be limited and may not reflect the model’s general applicability. To address this limitation, experimental data from three additional well-characterized cutinases (UniProt IDs: P52956, P41744, UPI00025213F3) were systematically curated from published literature~\cite{koschorreck2010heterologous, shirke2016toward, longhi1996dynamics, mannesse1995cutinase, longhi1999structure, egmond2000fusarium, araujo2007tailoring, chen2013cutinase, herrero2013surface}. As illustrated in Fig.~\ref{fig_application}f, mutations were annotated as thermally favorable (red squares) or detrimental (blue squares), with favorable mutations conferring enhanced temperature tolerance compared to WT. For cutinase P52956, segment transformer accurately predicted 12 out of 19 mutations (63.2\% accuracy). For P41744, all four mutations were correctly classified, while for UPI00025213F3, 8 of 10 mutations (80\% accuracy) were successfully predicted. These results demonstrate segment transformer’s potential to streamline thermostability engineering by reducing experimental screening efforts.

\section{Conclusion}

In this study, we presented the \textit{Segment Transformer}, a novel deep learning framework for enzyme temperature stability modeling. Designed with domain-specific architectural principles, the model accurately predicts temperature stability values from primary protein sequences. Our results show that Segment Transformer sets a new state-of-the-art in enzyme thermostability prediction, delivering both high accuracy and strong generalizability across diverse application scenarios.

The core innovation of Segment Transformer lies in its segment-level representation, which incorporates biological insights suggesting that different regions of an enzyme sequence contribute unequally to thermal behavior. By transforming amino acid-level features into segment-level representations and processing them with the Dual Grouped Segment Attention (DGSA) mechanism, Segment Transformer captures both local motifs and global dependencies. Its hierarchical structure is further enhanced by attention-based pooling and a multi-scale prediction head, which together yield robust and interpretable predictions. Evaluation on benchmark datasets confirmed the model’s strength, with an RMSE of 24.03, MAE of 18.09, and Pearson and Spearman correlations of 0.33 each.

Beyond accuracy, Segment Transformer includes several functional modules that enhance its practical utility. These include confidence intervals for predicted stability, region-wise importance scores, and interpretable temperature sensitivity maps. In a real-world enzyme engineering task, Segment Transformer guided the design of a thermostability-enhanced variant of cutinase, achieving a 1.64-fold increase in relative activity after heat treatment with only 17 mutations and no loss of native activity. Additionally, the model showed promising predictive accuracy for three other enzymes in independent mutation effect studies. These findings highlight its potential to accelerate enzyme engineering workflows focused on thermal property optimization.

Despite its strong performance, Segment Transformer has certain limitations. First, its predictions of mutation effects are not yet fully reliable. In the case of cutinase engineering, several predicted mutations resulted in decreased activity. This limitation stems from three key factors: (1) the model focuses solely on thermostability and does not jointly consider enzymatic activity, making some mutations functionally detrimental; (2) the absence of large-scale, annotated mutation-specific thermostability datasets restricts the model to a zero-shot setting for mutation prediction; and (3) while segment-level features improve prediction robustness, they may reduce sensitivity to fine-grained, single-residue changes, limiting resolution in mutation effect predictions.

Future work should aim to address these limitations by: (i) developing models that jointly predict both functional activity and thermal properties; (ii) curating large-scale, experimentally annotated datasets of mutation effects; and (iii) integrating hybrid features that combine both segment-level and residue-level information to improve mutation-level precision. With these advancements, Segment Transformer has the potential to become a general-purpose tool for rational enzyme design and stability optimization.

\begin{acknowledgement}

We acknowledge financial support from the National Natural Science Foundation of China (62176105), the China Scholarship Council (202406790100), Basic Research Funds for Central Universities in 2022-Zhishan Young Scholars Program (JUSRP622016), the Taishan Industrial Experts Program (NO. tscx202306145), the Yellow River Delta Industrial Leading Talent Program (NO. DYRC20200205), and the China Postdoctoral Science Foundation (NO. 2023M731167).

\textbf{Notes:} The authors declare no competing financial interests.

\end{acknowledgement}

\begin{suppinfo}

\begin{itemize}
  \item Data Availability: All data supporting the key findings of this study are available within the article and its Supplementary Information files. The data analyzed in this work are publicly accessible from the BRENDA database (\url{https://www.brenda-enzymes.org/}) and UniProt database (\url{https://www.uniprot.org/}). Source data are provided in Zenodo at \url{https://doi.org/10.5281/zenodo.15851711}.
  \item Code Availability: To support reproducibility and further research, all source code and detailed usage instructions are publicly available in our GitHub repository: \url{https://github.com/RIA-lab/Segment-transformer/tree/main}. The repository also includes an example script for predicting enzyme thermal stability. Model weights are available in Zenodo at \url{https://doi.org/10.5281/zenodo.15851719}.
\end{itemize}

\end{suppinfo}

\bibliography{achemso-demo}

\providecommand{\latin}[1]{#1}
\makeatletter
\providecommand{\doi}
  {\begingroup\let\do\@makeother\dospecials
  \catcode`\{=1 \catcode`\}=2 \doi@aux}
\providecommand{\doi@aux}[1]{\endgroup\texttt{#1}}
\makeatother
\providecommand*\mcitethebibliography{\thebibliography}
\csname @ifundefined\endcsname{endmcitethebibliography}  {\let\endmcitethebibliography\endthebibliography}{}
\begin{mcitethebibliography}{49}
\providecommand*\natexlab[1]{#1}
\providecommand*\mciteSetBstSublistMode[1]{}
\providecommand*\mciteSetBstMaxWidthForm[2]{}
\providecommand*\mciteBstWouldAddEndPuncttrue
  {\def\EndOfBibitem{\unskip.}}
\providecommand*\mciteBstWouldAddEndPunctfalse
  {\let\EndOfBibitem\relax}
\providecommand*\mciteSetBstMidEndSepPunct[3]{}
\providecommand*\mciteSetBstSublistLabelBeginEnd[3]{}
\providecommand*\EndOfBibitem{}
\mciteSetBstSublistMode{f}
\mciteSetBstMaxWidthForm{subitem}{(\alph{mcitesubitemcount})}
\mciteSetBstSublistLabelBeginEnd
  {\mcitemaxwidthsubitemform\space}
  {\relax}
  {\relax}

\bibitem[Daniel \latin{et~al.}(2001)Daniel, Danson, and Eisenthal]{daniel2001temperature}
Daniel,~R.~M.; Danson,~M.~J.; Eisenthal,~R. The temperature optima of enzymes: a new perspective on an old phenomenon. \emph{Trends in biochemical sciences} \textbf{2001}, \emph{26}, 223--225\relax
\mciteBstWouldAddEndPuncttrue
\mciteSetBstMidEndSepPunct{\mcitedefaultmidpunct}
{\mcitedefaultendpunct}{\mcitedefaultseppunct}\relax
\EndOfBibitem
\bibitem[Daniel \latin{et~al.}(2008)Daniel, Danson, Eisenthal, Lee, and Peterson]{daniel2008effect}
Daniel,~R.~M.; Danson,~M.~J.; Eisenthal,~R.; Lee,~C.~K.; Peterson,~M.~E. The effect of temperature on enzyme activity: new insights and their implications. \emph{Extremophiles} \textbf{2008}, \emph{12}, 51--59\relax
\mciteBstWouldAddEndPuncttrue
\mciteSetBstMidEndSepPunct{\mcitedefaultmidpunct}
{\mcitedefaultendpunct}{\mcitedefaultseppunct}\relax
\EndOfBibitem
\bibitem[Daniel and Danson(2010)Daniel, and Danson]{daniel2010new}
Daniel,~R.~M.; Danson,~M.~J. A new understanding of how temperature affects the catalytic activity of enzymes. \emph{Trends in biochemical sciences} \textbf{2010}, \emph{35}, 584--591\relax
\mciteBstWouldAddEndPuncttrue
\mciteSetBstMidEndSepPunct{\mcitedefaultmidpunct}
{\mcitedefaultendpunct}{\mcitedefaultseppunct}\relax
\EndOfBibitem
\bibitem[Lee \latin{et~al.}(2013)Lee, Monk, and Daniel]{lee2013determination}
Lee,~C.~K.; Monk,~C.~R.; Daniel,~R.~M. Determination of enzyme thermal parameters for rational enzyme engineering and environmental/evolutionary studies. \emph{Protein Nanotechnology: Protocols, Instrumentation, and Applications, Second Edition} \textbf{2013}, 219--230\relax
\mciteBstWouldAddEndPuncttrue
\mciteSetBstMidEndSepPunct{\mcitedefaultmidpunct}
{\mcitedefaultendpunct}{\mcitedefaultseppunct}\relax
\EndOfBibitem
\bibitem[Xu \latin{et~al.}(2020)Xu, Cen, Zou, Xue, and Zheng]{xu2020recent}
Xu,~Z.; Cen,~Y.-K.; Zou,~S.-P.; Xue,~Y.-P.; Zheng,~Y.-G. Recent advances in the improvement of enzyme thermostability by structure modification. \emph{Critical reviews in biotechnology} \textbf{2020}, \emph{40}, 83--98\relax
\mciteBstWouldAddEndPuncttrue
\mciteSetBstMidEndSepPunct{\mcitedefaultmidpunct}
{\mcitedefaultendpunct}{\mcitedefaultseppunct}\relax
\EndOfBibitem
\bibitem[Consortium(2024)]{10.1093/nar/gkae1010}
Consortium,~T.~U. UniProt: the Universal Protein Knowledgebase in 2025. \emph{Nucleic Acids Research} \textbf{2024}, \emph{53}, D609--D617\relax
\mciteBstWouldAddEndPuncttrue
\mciteSetBstMidEndSepPunct{\mcitedefaultmidpunct}
{\mcitedefaultendpunct}{\mcitedefaultseppunct}\relax
\EndOfBibitem
\bibitem[Chang \latin{et~al.}(2020)Chang, Jeske, Ulbrich, Hofmann, Koblitz, Schomburg, Neumann-Schaal, Jahn, and Schomburg]{10.1093/nar/gkaa1025}
Chang,~A.; Jeske,~L.; Ulbrich,~S.; Hofmann,~J.; Koblitz,~J.; Schomburg,~I.; Neumann-Schaal,~M.; Jahn,~D.; Schomburg,~D. BRENDA, the ELIXIR core data resource in 2021: new developments and updates. \emph{Nucleic Acids Research} \textbf{2020}, \emph{49}, D498--D508\relax
\mciteBstWouldAddEndPuncttrue
\mciteSetBstMidEndSepPunct{\mcitedefaultmidpunct}
{\mcitedefaultendpunct}{\mcitedefaultseppunct}\relax
\EndOfBibitem
\bibitem[Yan and Wu(2012)Yan, and Wu]{yan2012prediction}
Yan,~S.-M.; Wu,~G. Prediction of optimal pH and temperature of cellulases using neural network. \emph{Protein and Peptide Letters} \textbf{2012}, \emph{19}, 29--39\relax
\mciteBstWouldAddEndPuncttrue
\mciteSetBstMidEndSepPunct{\mcitedefaultmidpunct}
{\mcitedefaultendpunct}{\mcitedefaultseppunct}\relax
\EndOfBibitem
\bibitem[Yan and Wu(2019)Yan, and Wu]{yan2019predictors}
Yan,~S.; Wu,~G. Predictors for predicting temperature optimum in beta-glucosidases. \emph{Journal of Biomedical Science and Engineering} \textbf{2019}, \emph{12}, 414--426\relax
\mciteBstWouldAddEndPuncttrue
\mciteSetBstMidEndSepPunct{\mcitedefaultmidpunct}
{\mcitedefaultendpunct}{\mcitedefaultseppunct}\relax
\EndOfBibitem
\bibitem[Zhang and Ge(2012)Zhang, and Ge]{zhang2012prediction}
Zhang,~G.; Ge,~H. Prediction of xylanase optimal temperature by support vector regression. \emph{Electronic Journal of Biotechnology} \textbf{2012}, \emph{15}, 7--7\relax
\mciteBstWouldAddEndPuncttrue
\mciteSetBstMidEndSepPunct{\mcitedefaultmidpunct}
{\mcitedefaultendpunct}{\mcitedefaultseppunct}\relax
\EndOfBibitem
\bibitem[Chu \latin{et~al.}(2016)Chu, Yi, Zeng, and Zhang]{chu2016predicting}
Chu,~Y.; Yi,~Z.; Zeng,~R.; Zhang,~G. Predicting the optimum temperature of $\beta$-agarase based on the relative solvent accessibility of amino acids. \emph{Journal of Molecular Catalysis B: Enzymatic} \textbf{2016}, \emph{129}, 47--53\relax
\mciteBstWouldAddEndPuncttrue
\mciteSetBstMidEndSepPunct{\mcitedefaultmidpunct}
{\mcitedefaultendpunct}{\mcitedefaultseppunct}\relax
\EndOfBibitem
\bibitem[Foroozandeh~Shahraki \latin{et~al.}(2021)Foroozandeh~Shahraki, Farhadyar, Kavousi, Azarabad, Boroomand, Ariaeenejad, and Hosseini~Salekdeh]{foroozandeh2021generalized}
Foroozandeh~Shahraki,~M.; Farhadyar,~K.; Kavousi,~K.; Azarabad,~M.~H.; Boroomand,~A.; Ariaeenejad,~S.; Hosseini~Salekdeh,~G. A generalized machine-learning aided method for targeted identification of industrial enzymes from metagenome: A xylanase temperature dependence case study. \emph{Biotechnology and Bioengineering} \textbf{2021}, \emph{118}, 759--769\relax
\mciteBstWouldAddEndPuncttrue
\mciteSetBstMidEndSepPunct{\mcitedefaultmidpunct}
{\mcitedefaultendpunct}{\mcitedefaultseppunct}\relax
\EndOfBibitem
\bibitem[Li \latin{et~al.}(2019)Li, Rabe, Nielsen, and Engqvist]{li2019machine}
Li,~G.; Rabe,~K.~S.; Nielsen,~J.; Engqvist,~M.~K. Machine learning applied to predicting microorganism growth temperatures and enzyme catalytic optima. \emph{ACS synthetic biology} \textbf{2019}, \emph{8}, 1411--1420\relax
\mciteBstWouldAddEndPuncttrue
\mciteSetBstMidEndSepPunct{\mcitedefaultmidpunct}
{\mcitedefaultendpunct}{\mcitedefaultseppunct}\relax
\EndOfBibitem
\bibitem[Gado \latin{et~al.}(2020)Gado, Beckham, and Payne]{gado2020improving}
Gado,~J.~E.; Beckham,~G.~T.; Payne,~C.~M. Improving enzyme optimum temperature prediction with resampling strategies and ensemble learning. \emph{Journal of Chemical Information and Modeling} \textbf{2020}, \emph{60}, 4098--4107\relax
\mciteBstWouldAddEndPuncttrue
\mciteSetBstMidEndSepPunct{\mcitedefaultmidpunct}
{\mcitedefaultendpunct}{\mcitedefaultseppunct}\relax
\EndOfBibitem
\bibitem[Wang \latin{et~al.}(2024)Wang, Zong, Zhou, Xu, He, and Quan]{wang2024artificial}
Wang,~X.; Zong,~Y.; Zhou,~X.; Xu,~L.; He,~W.; Quan,~S. Artificial Intelligence-Powered Construction of a Microbial Optimal Growth Temperature Database and Its Impact on Enzyme Optimal Temperature Prediction. \emph{The Journal of Physical Chemistry B} \textbf{2024}, \emph{128}, 2281--2292\relax
\mciteBstWouldAddEndPuncttrue
\mciteSetBstMidEndSepPunct{\mcitedefaultmidpunct}
{\mcitedefaultendpunct}{\mcitedefaultseppunct}\relax
\EndOfBibitem
\bibitem[Li \latin{et~al.}(2022)Li, Buric, Zrimec, Viknander, Nielsen, Zelezniak, and Engqvist]{li2022learning}
Li,~G.; Buric,~F.; Zrimec,~J.; Viknander,~S.; Nielsen,~J.; Zelezniak,~A.; Engqvist,~M.~K. Learning deep representations of enzyme thermal adaptation. \emph{Protein Science} \textbf{2022}, \emph{31}, e4480\relax
\mciteBstWouldAddEndPuncttrue
\mciteSetBstMidEndSepPunct{\mcitedefaultmidpunct}
{\mcitedefaultendpunct}{\mcitedefaultseppunct}\relax
\EndOfBibitem
\bibitem[Zhang \latin{et~al.}(2022)Zhang, Guan, Xu, Liu, Zhang, Sun, Yao, Huang, Wu, and Tian]{zhang2022novel}
Zhang,~Y.; Guan,~F.; Xu,~G.; Liu,~X.; Zhang,~Y.; Sun,~J.; Yao,~B.; Huang,~H.; Wu,~N.; Tian,~J. A novel thermophilic chitinase directly mined from the marine metagenome using the deep learning tool Preoptem. \emph{Bioresources and Bioprocessing} \textbf{2022}, \emph{9}, 54\relax
\mciteBstWouldAddEndPuncttrue
\mciteSetBstMidEndSepPunct{\mcitedefaultmidpunct}
{\mcitedefaultendpunct}{\mcitedefaultseppunct}\relax
\EndOfBibitem
\bibitem[Qiu \latin{et~al.}(2024)Qiu, Hu, Zhao, Xu, and Yang]{qiu2024seq2topt}
Qiu,~S.; Hu,~B.; Zhao,~J.; Xu,~W.; Yang,~A. Seq2Topt: a sequence-based deep learning predictor of enzyme optimal temperature. \emph{bioRxiv} \textbf{2024}, 2024--08\relax
\mciteBstWouldAddEndPuncttrue
\mciteSetBstMidEndSepPunct{\mcitedefaultmidpunct}
{\mcitedefaultendpunct}{\mcitedefaultseppunct}\relax
\EndOfBibitem
\bibitem[Elnaggar \latin{et~al.}(2021)Elnaggar, Heinzinger, Dallago, Rehawi, Yu, Jones, Gibbs, Feher, Angerer, Steinegger, Bhowmik, and Rost]{9477085}
Elnaggar,~A.; Heinzinger,~M.; Dallago,~C.; Rehawi,~G.; Yu,~W.; Jones,~L.; Gibbs,~T.; Feher,~T.; Angerer,~C.; Steinegger,~M.; Bhowmik,~D.; Rost,~B. ProtTrans: Towards Cracking the Language of Lifes Code Through Self-Supervised Deep Learning and High Performance Computing. \emph{IEEE Transactions on Pattern Analysis and Machine Intelligence} \textbf{2021}, 1--1\relax
\mciteBstWouldAddEndPuncttrue
\mciteSetBstMidEndSepPunct{\mcitedefaultmidpunct}
{\mcitedefaultendpunct}{\mcitedefaultseppunct}\relax
\EndOfBibitem
\bibitem[Lin \latin{et~al.}(2022)Lin, Akin, Rao, Hie, Zhu, Lu, Smetanin, dos Santos~Costa, Fazel-Zarandi, Sercu, Candido, \latin{et~al.} others]{lin2022language}
Lin,~Z.; Akin,~H.; Rao,~R.; Hie,~B.; Zhu,~Z.; Lu,~W.; Smetanin,~N.; dos Santos~Costa,~A.; Fazel-Zarandi,~M.; Sercu,~T.; Candido,~S.; others Language models of protein sequences at the scale of evolution enable accurate structure prediction. \emph{bioRxiv} \textbf{2022}, \relax
\mciteBstWouldAddEndPunctfalse
\mciteSetBstMidEndSepPunct{\mcitedefaultmidpunct}
{}{\mcitedefaultseppunct}\relax
\EndOfBibitem
\bibitem[Chronopoulou \latin{et~al.}(2023)Chronopoulou, Mutabdzija, Poudel, Papageorgiou, and Labrou]{chronopoulou2023key}
Chronopoulou,~E.~G.; Mutabdzija,~L.; Poudel,~N.; Papageorgiou,~A.~C.; Labrou,~N.~E. A Key Role in Catalysis and Enzyme Thermostability of a Conserved Helix H5 Motif of Human Glutathione Transferase A1-1. \emph{International journal of molecular sciences} \textbf{2023}, \emph{24}, 3700\relax
\mciteBstWouldAddEndPuncttrue
\mciteSetBstMidEndSepPunct{\mcitedefaultmidpunct}
{\mcitedefaultendpunct}{\mcitedefaultseppunct}\relax
\EndOfBibitem
\bibitem[Lee \latin{et~al.}(2014)Lee, Wang, Hwang, and Tseng]{lee2014protein}
Lee,~C.-W.; Wang,~H.-J.; Hwang,~J.-K.; Tseng,~C.-P. Protein thermal stability enhancement by designing salt bridges: a combined computational and experimental study. \emph{PloS one} \textbf{2014}, \emph{9}, e112751\relax
\mciteBstWouldAddEndPuncttrue
\mciteSetBstMidEndSepPunct{\mcitedefaultmidpunct}
{\mcitedefaultendpunct}{\mcitedefaultseppunct}\relax
\EndOfBibitem
\bibitem[Kannan and Vishveshwara(2000)Kannan, and Vishveshwara]{kannan2000aromatic}
Kannan,~N.; Vishveshwara,~S. Aromatic clusters: a determinant of thermal stability of thermophilic proteins. \emph{Protein engineering} \textbf{2000}, \emph{13}, 753--761\relax
\mciteBstWouldAddEndPuncttrue
\mciteSetBstMidEndSepPunct{\mcitedefaultmidpunct}
{\mcitedefaultendpunct}{\mcitedefaultseppunct}\relax
\EndOfBibitem
\bibitem[Musil \latin{et~al.}(2017)Musil, Stourac, Bendl, Brezovsky, Prokop, Zendulka, Martinek, Bednar, and Damborsky]{musil2017fireprot}
Musil,~M.; Stourac,~J.; Bendl,~J.; Brezovsky,~J.; Prokop,~Z.; Zendulka,~J.; Martinek,~T.; Bednar,~D.; Damborsky,~J. FireProt: web server for automated design of thermostable proteins. \emph{Nucleic acids research} \textbf{2017}, \emph{45}, W393--W399\relax
\mciteBstWouldAddEndPuncttrue
\mciteSetBstMidEndSepPunct{\mcitedefaultmidpunct}
{\mcitedefaultendpunct}{\mcitedefaultseppunct}\relax
\EndOfBibitem
\bibitem[Liu and Kuhlman(2006)Liu, and Kuhlman]{liu2006rosettadesign}
Liu,~Y.; Kuhlman,~B. RosettaDesign server for protein design. \emph{Nucleic acids research} \textbf{2006}, \emph{34}, W235--W238\relax
\mciteBstWouldAddEndPuncttrue
\mciteSetBstMidEndSepPunct{\mcitedefaultmidpunct}
{\mcitedefaultendpunct}{\mcitedefaultseppunct}\relax
\EndOfBibitem
\bibitem[Goldenzweig \latin{et~al.}(2016)Goldenzweig, Goldsmith, Hill, Gertman, Laurino, Ashani, Dym, Unger, Albeck, Prilusky, \latin{et~al.} others]{goldenzweig2016automated}
Goldenzweig,~A.; Goldsmith,~M.; Hill,~S.~E.; Gertman,~O.; Laurino,~P.; Ashani,~Y.; Dym,~O.; Unger,~T.; Albeck,~S.; Prilusky,~J.; others Automated structure-and sequence-based design of proteins for high bacterial expression and stability. \emph{Molecular cell} \textbf{2016}, \emph{63}, 337--346\relax
\mciteBstWouldAddEndPuncttrue
\mciteSetBstMidEndSepPunct{\mcitedefaultmidpunct}
{\mcitedefaultendpunct}{\mcitedefaultseppunct}\relax
\EndOfBibitem
\bibitem[Gado \latin{et~al.}(2025)Gado, Knotts, Shaw, Marks, Gauthier, Sander, and Beckham]{gado2025machine}
Gado,~J.~E.; Knotts,~M.; Shaw,~A.~Y.; Marks,~D.; Gauthier,~N.~P.; Sander,~C.; Beckham,~G.~T. Machine learning prediction of enzyme optimum pH. \emph{Nature Machine Intelligence} \textbf{2025}, 1--14\relax
\mciteBstWouldAddEndPuncttrue
\mciteSetBstMidEndSepPunct{\mcitedefaultmidpunct}
{\mcitedefaultendpunct}{\mcitedefaultseppunct}\relax
\EndOfBibitem
\bibitem[Steinegger and S{\"o}ding(2017)Steinegger, and S{\"o}ding]{steinegger2017mmseqs2}
Steinegger,~M.; S{\"o}ding,~J. MMseqs2 enables sensitive protein sequence searching for the analysis of massive data sets. \emph{Nature biotechnology} \textbf{2017}, \emph{35}, 1026--1028\relax
\mciteBstWouldAddEndPuncttrue
\mciteSetBstMidEndSepPunct{\mcitedefaultmidpunct}
{\mcitedefaultendpunct}{\mcitedefaultseppunct}\relax
\EndOfBibitem
\bibitem[Lin \latin{et~al.}(2023)Lin, Akin, Rao, Hie, Zhu, Lu, Smetanin, Verkuil, Kabeli, Shmueli, \latin{et~al.} others]{lin2023evolutionary}
Lin,~Z.; Akin,~H.; Rao,~R.; Hie,~B.; Zhu,~Z.; Lu,~W.; Smetanin,~N.; Verkuil,~R.; Kabeli,~O.; Shmueli,~Y.; others Evolutionary-scale prediction of atomic-level protein structure with a language model. \emph{Science} \textbf{2023}, \emph{379}, 1123--1130\relax
\mciteBstWouldAddEndPuncttrue
\mciteSetBstMidEndSepPunct{\mcitedefaultmidpunct}
{\mcitedefaultendpunct}{\mcitedefaultseppunct}\relax
\EndOfBibitem
\bibitem[van~der Maaten and Hinton(2008)van~der Maaten, and Hinton]{JMLR:v9:vandermaaten08a}
van~der Maaten,~L.; Hinton,~G. Visualizing Data using t-SNE. \emph{Journal of Machine Learning Research} \textbf{2008}, \emph{9}, 2579--2605\relax
\mciteBstWouldAddEndPuncttrue
\mciteSetBstMidEndSepPunct{\mcitedefaultmidpunct}
{\mcitedefaultendpunct}{\mcitedefaultseppunct}\relax
\EndOfBibitem
\bibitem[Rao \latin{et~al.}(2022)Rao, Wang, Huo, Su, Guo, Yang, Wei, Tao, Chen, and Wu]{rao2022trehalose}
Rao,~D.; Wang,~L.; Huo,~R.; Su,~L.; Guo,~Z.; Yang,~W.; Wei,~B.; Tao,~X.; Chen,~S.; Wu,~J. Trehalose promotes high-level heterologous expression of 4, 6-$\alpha$-glucanotransferase GtfR2 in Escherichia coli and mechanistic analysis. \emph{International Journal of Biological Macromolecules} \textbf{2022}, \emph{210}, 315--323\relax
\mciteBstWouldAddEndPuncttrue
\mciteSetBstMidEndSepPunct{\mcitedefaultmidpunct}
{\mcitedefaultendpunct}{\mcitedefaultseppunct}\relax
\EndOfBibitem
\bibitem[Roe and Cheatham~III(2013)Roe, and Cheatham~III]{roe2013ptraj}
Roe,~D.~R.; Cheatham~III,~T.~E. PTRAJ and CPPTRAJ: software for processing and analysis of molecular dynamics trajectory data. \emph{Journal of chemical theory and computation} \textbf{2013}, \emph{9}, 3084--3095\relax
\mciteBstWouldAddEndPuncttrue
\mciteSetBstMidEndSepPunct{\mcitedefaultmidpunct}
{\mcitedefaultendpunct}{\mcitedefaultseppunct}\relax
\EndOfBibitem
\bibitem[Schuster and Paliwal(1997)Schuster, and Paliwal]{schuster1997bidirectional}
Schuster,~M.; Paliwal,~K.~K. Bidirectional recurrent neural networks. \emph{IEEE transactions on Signal Processing} \textbf{1997}, \emph{45}, 2673--2681\relax
\mciteBstWouldAddEndPuncttrue
\mciteSetBstMidEndSepPunct{\mcitedefaultmidpunct}
{\mcitedefaultendpunct}{\mcitedefaultseppunct}\relax
\EndOfBibitem
\bibitem[Collobert \latin{et~al.}(2011)Collobert, Weston, Bottou, Karlen, Kavukcuoglu, and Kuksa]{collobert2011natural}
Collobert,~R.; Weston,~J.; Bottou,~L.; Karlen,~M.; Kavukcuoglu,~K.; Kuksa,~P. Natural language processing (almost) from scratch. \textbf{2011}, \relax
\mciteBstWouldAddEndPunctfalse
\mciteSetBstMidEndSepPunct{\mcitedefaultmidpunct}
{}{\mcitedefaultseppunct}\relax
\EndOfBibitem
\bibitem[Vaswani \latin{et~al.}(2017)Vaswani, Shazeer, Parmar, Uszkoreit, Jones, Gomez, Kaiser, and Polosukhin]{vaswani2017attention}
Vaswani,~A.; Shazeer,~N.; Parmar,~N.; Uszkoreit,~J.; Jones,~L.; Gomez,~A.~N.; Kaiser,~{\L}.; Polosukhin,~I. Attention is all you need. \emph{Advances in neural information processing systems} \textbf{2017}, \emph{30}\relax
\mciteBstWouldAddEndPuncttrue
\mciteSetBstMidEndSepPunct{\mcitedefaultmidpunct}
{\mcitedefaultendpunct}{\mcitedefaultseppunct}\relax
\EndOfBibitem
\bibitem[St{\"a}rk \latin{et~al.}(2021)St{\"a}rk, Dallago, Heinzinger, and Rost]{stark2021light}
St{\"a}rk,~H.; Dallago,~C.; Heinzinger,~M.; Rost,~B. Light attention predicts protein location from the language of life. \emph{Bioinformatics Advances} \textbf{2021}, \emph{1}, vbab035\relax
\mciteBstWouldAddEndPuncttrue
\mciteSetBstMidEndSepPunct{\mcitedefaultmidpunct}
{\mcitedefaultendpunct}{\mcitedefaultseppunct}\relax
\EndOfBibitem
\bibitem[Elman(1990)]{elman1990finding}
Elman,~J.~L. Finding structure in time. \emph{Cognitive science} \textbf{1990}, \emph{14}, 179--211\relax
\mciteBstWouldAddEndPuncttrue
\mciteSetBstMidEndSepPunct{\mcitedefaultmidpunct}
{\mcitedefaultendpunct}{\mcitedefaultseppunct}\relax
\EndOfBibitem
\bibitem[Pud{\v{z}}iuvelyt{\.e} \latin{et~al.}(2024)Pud{\v{z}}iuvelyt{\.e}, Olechnovi{\v{c}}, Godliauskaite, Sermokas, Urbaitis, Gasiunas, and Kazlauskas]{pudvziuvelyte2024temstapro}
Pud{\v{z}}iuvelyt{\.e},~I.; Olechnovi{\v{c}},~K.; Godliauskaite,~E.; Sermokas,~K.; Urbaitis,~T.; Gasiunas,~G.; Kazlauskas,~D. TemStaPro: protein thermostability prediction using sequence representations from protein language models. \emph{Bioinformatics} \textbf{2024}, \emph{40}, btae157\relax
\mciteBstWouldAddEndPuncttrue
\mciteSetBstMidEndSepPunct{\mcitedefaultmidpunct}
{\mcitedefaultendpunct}{\mcitedefaultseppunct}\relax
\EndOfBibitem
\bibitem[Qiu \latin{et~al.}(2025)Qiu, Hu, Zhao, Xu, and Yang]{qiu2025seq2topt}
Qiu,~S.; Hu,~B.; Zhao,~J.; Xu,~W.; Yang,~A. Seq2Topt: a sequence-based deep learning predictor of enzyme optimal temperature. \emph{Briefings in Bioinformatics} \textbf{2025}, \emph{26}, bbaf114\relax
\mciteBstWouldAddEndPuncttrue
\mciteSetBstMidEndSepPunct{\mcitedefaultmidpunct}
{\mcitedefaultendpunct}{\mcitedefaultseppunct}\relax
\EndOfBibitem
\bibitem[Koschorreck \latin{et~al.}(2010)Koschorreck, Liu, Kazenwadel, Schmid, and Hauer]{koschorreck2010heterologous}
Koschorreck,~K.; Liu,~D.; Kazenwadel,~C.; Schmid,~R.~D.; Hauer,~B. Heterologous expression, characterization and site-directed mutagenesis of cutinase CUTAB1 from Alternaria brassicicola. \emph{Applied microbiology and biotechnology} \textbf{2010}, \emph{87}, 991--997\relax
\mciteBstWouldAddEndPuncttrue
\mciteSetBstMidEndSepPunct{\mcitedefaultmidpunct}
{\mcitedefaultendpunct}{\mcitedefaultseppunct}\relax
\EndOfBibitem
\bibitem[Shirke \latin{et~al.}(2016)Shirke, Basore, Butterfoss, Bonneau, Bystroff, and Gross]{shirke2016toward}
Shirke,~A.~N.; Basore,~D.; Butterfoss,~G.~L.; Bonneau,~R.; Bystroff,~C.; Gross,~R.~A. Toward rational thermostabilization of Aspergillus oryzae cutinase: insights into catalytic and structural stability. \emph{Proteins: Structure, Function, and Bioinformatics} \textbf{2016}, \emph{84}, 60--72\relax
\mciteBstWouldAddEndPuncttrue
\mciteSetBstMidEndSepPunct{\mcitedefaultmidpunct}
{\mcitedefaultendpunct}{\mcitedefaultseppunct}\relax
\EndOfBibitem
\bibitem[Longhi \latin{et~al.}(1996)Longhi, Nicolas, Creveld, Egmond, Verrips, de~Vlieg, Martinez, and Cambillau]{longhi1996dynamics}
Longhi,~S.; Nicolas,~A.; Creveld,~L.; Egmond,~M.; Verrips,~C.~T.; de~Vlieg,~J.; Martinez,~C.; Cambillau,~C. Dynamics of Fusarium solani cutinase investigated through structural comparison among different crystal forms of its variants. \emph{Proteins: Structure, Function, and Bioinformatics} \textbf{1996}, \emph{26}, 442--458\relax
\mciteBstWouldAddEndPuncttrue
\mciteSetBstMidEndSepPunct{\mcitedefaultmidpunct}
{\mcitedefaultendpunct}{\mcitedefaultseppunct}\relax
\EndOfBibitem
\bibitem[Mannesse \latin{et~al.}(1995)Mannesse, Cox, Koops, Verheij, de~Haas, Egmond, van~der Hijden, and de~Vlieg]{mannesse1995cutinase}
Mannesse,~M.~L.; Cox,~R.~C.; Koops,~B.~C.; Verheij,~H.~M.; de~Haas,~G.~H.; Egmond,~M.~R.; van~der Hijden,~H.~T.; de~Vlieg,~J. Cutinase from Fusarium solani pisi hydrolyzing triglyceride analogs. Effect of acyl chain length and position in the substrate molecule on activity and enantioselectivity. \emph{Biochemistry} \textbf{1995}, \emph{34}, 6400--6407\relax
\mciteBstWouldAddEndPuncttrue
\mciteSetBstMidEndSepPunct{\mcitedefaultmidpunct}
{\mcitedefaultendpunct}{\mcitedefaultseppunct}\relax
\EndOfBibitem
\bibitem[Longhi and Cambillau(1999)Longhi, and Cambillau]{longhi1999structure}
Longhi,~S.; Cambillau,~C. Structure-activity of cutinase, a small lipolytic enzyme. \emph{Biochimica et Biophysica Acta (BBA)-Molecular and Cell Biology of Lipids} \textbf{1999}, \emph{1441}, 185--196\relax
\mciteBstWouldAddEndPuncttrue
\mciteSetBstMidEndSepPunct{\mcitedefaultmidpunct}
{\mcitedefaultendpunct}{\mcitedefaultseppunct}\relax
\EndOfBibitem
\bibitem[Egmond and de~Vlieg(2000)Egmond, and de~Vlieg]{egmond2000fusarium}
Egmond,~M.~R.; de~Vlieg,~J. Fusarium solani pisi cutinase. \emph{Biochimie} \textbf{2000}, \emph{82}, 1015--1021\relax
\mciteBstWouldAddEndPuncttrue
\mciteSetBstMidEndSepPunct{\mcitedefaultmidpunct}
{\mcitedefaultendpunct}{\mcitedefaultseppunct}\relax
\EndOfBibitem
\bibitem[Ara{\'u}jo \latin{et~al.}(2007)Ara{\'u}jo, Silva, O’Neill, Micaelo, Guebitz, Soares, Casal, and Cavaco-Paulo]{araujo2007tailoring}
Ara{\'u}jo,~R.; Silva,~C.; O’Neill,~A.; Micaelo,~N.; Guebitz,~G.; Soares,~C.~M.; Casal,~M.; Cavaco-Paulo,~A. Tailoring cutinase activity towards polyethylene terephthalate and polyamide 6, 6 fibers. \emph{Journal of Biotechnology} \textbf{2007}, \emph{128}, 849--857\relax
\mciteBstWouldAddEndPuncttrue
\mciteSetBstMidEndSepPunct{\mcitedefaultmidpunct}
{\mcitedefaultendpunct}{\mcitedefaultseppunct}\relax
\EndOfBibitem
\bibitem[Chen \latin{et~al.}(2013)Chen, Su, Chen, and Wu]{chen2013cutinase}
Chen,~S.; Su,~L.; Chen,~J.; Wu,~J. Cutinase: characteristics, preparation, and application. \emph{Biotechnology advances} \textbf{2013}, \emph{31}, 1754--1767\relax
\mciteBstWouldAddEndPuncttrue
\mciteSetBstMidEndSepPunct{\mcitedefaultmidpunct}
{\mcitedefaultendpunct}{\mcitedefaultseppunct}\relax
\EndOfBibitem
\bibitem[Herrero~Acero \latin{et~al.}(2013)Herrero~Acero, Ribitsch, Dellacher, Zitzenbacher, Marold, Steinkellner, Gruber, Schwab, and Guebitz]{herrero2013surface}
Herrero~Acero,~E.; Ribitsch,~D.; Dellacher,~A.; Zitzenbacher,~S.; Marold,~A.; Steinkellner,~G.; Gruber,~K.; Schwab,~H.; Guebitz,~G.~M. Surface engineering of a cutinase from Thermobifida cellulosilytica for improved polyester hydrolysis. \emph{Biotechnology and Bioengineering} \textbf{2013}, \emph{110}, 2581--2590\relax
\mciteBstWouldAddEndPuncttrue
\mciteSetBstMidEndSepPunct{\mcitedefaultmidpunct}
{\mcitedefaultendpunct}{\mcitedefaultseppunct}\relax
\EndOfBibitem
\end{mcitethebibliography}

\end{document}